\documentclass[a4paper,fleqn]{cas-dc}
\usepackage[numbers]{natbib}
\usepackage{amsmath,amsfonts}
\usepackage{algorithmic}
\usepackage{algorithm}
\usepackage{array}
\usepackage[caption=false,font=footnotesize,labelfont=rm,textfont=rm]{subfig}
\usepackage{textcomp}
\usepackage{stfloats}
\usepackage{url}
\usepackage{verbatim}
\usepackage{graphicx}
\usepackage{float}
\usepackage{bm}
\usepackage{multirow}
\usepackage{setspace}
\usepackage{threeparttable}
\usepackage{silence}
\usepackage{amssymb}
\usepackage{color}
\usepackage{xcolor}
\usepackage{pifont}
\usepackage{utfsym}
\usepackage{tabularx}
\usepackage{siunitx}
\usepackage{tabularx}
\usepackage{enumitem}
\DeclareSIUnit\diag{diag}
\def\tsc#1{\csdef{#1}{\textsc{\lowercase{#1}}\xspace}}
\tsc{WGM}
\tsc{QE}

\begin{document}
\let\WriteBookmarks\relax
\def\floatpagepagefraction{1}
\def\textpagefraction{.001}

\shorttitle{Adaptive Learning-based MPC Strategy for Drift Vehicles}    

\shortauthors{Bei Zhou, Cheng Hu et al.}  

\title [mode = title]{Adaptive Learning-based Model Predictive Control Strategy for Drift Vehicles}

\tnotetext[1]{Supported by Jianbing Lingyan Foundation of Zhejiang Province, P.R. China (Grant No. 2023C01022)} 

\author[1]{Bei Zhou}
\ead{zhoubei@zju.edu.cn}

\affiliation[1]{organization={the State Key Laboratory of Industrial Control Technology, Zhejiang University},
	addressline={38 Zheda Road, Xihu District}, 
	city={Hang zhou},
	postcode={310027}, 
	state={Zhejiang},
	country={China}}

\affiliation[2]{organization={Hybrid Robotics Group at the Department of Mechanical Engineering,
UC Berkeley},
	country={USA}}

\author[1]{Cheng Hu}
\ead{22032081@zju.edu.cn}

\author[2]{Jun Zeng}
\ead{zengjunsjtu@berkeley.edu}

\author[1]{Zhouheng Li}
\ead{zh.li@zju.edu.cn}

\author[3]{Johannes Betz}
\ead{johannes.betz@tum.de}

\affiliation[3]{organization={Professorship Autonomous Vehicle Systems and Munich and Munich Institute of Robotics and Machine Intelligence (MIRMI), Technical University of Munich, 85748 Garching},
country={Germany}}

\author[1]{Lei Xie}
\cormark[1]
\ead{lxie@iipc.zju.edu.cn}

\author[1]{Hongye Su}
\ead{hysu@iipc.zju.edu.cn}

\cortext[1]{Corresponding author}

\begin{abstract}
Drift vehicle control offers valuable insights to support safe autonomous driving in extreme conditions, which hinges on tracking a particular path while maintaining the vehicle states near the drift equilibrium points (DEP).
However, conventional tracking methods are not adaptable for drift vehicles due to their opposite steering angle and yaw rate.
In this paper, we propose an adaptive path tracking (APT) control method to dynamically adjust drift states to follow the reference path, improving the commonly utilized predictive path tracking methods with released computation burden.
Furthermore, existing control strategies necessitate a precise system model to calculate the DEP, which can be more intractable due to the highly nonlinear drift dynamics and sensitive vehicle parameters. 
To tackle this problem, an adaptive learning-based model predictive control (ALMPC) strategy is proposed based on the APT method, where an upper-level Bayesian optimization is employed to learn the DEP and APT control law to instruct a lower-level MPC drift controller.	
This hierarchical system architecture can also resolve the inherent control conflict between path tracking and drifting by separating these objectives into different layers.
The ALMPC strategy is verified on the Matlab-Carsim platform, and simulation results demonstrate its effectiveness in controlling the drift vehicle to follow a clothoid-based reference path even with the misidentified road friction parameter.
\end{abstract}


\maketitle

\section{Introduction}
The rising popularity of autonomous driving has led to a new research field in autonomous racing, where vehicles are pushed to their limits in high-speed, dynamic, and uncertain environments \cite{AutonomousVehicles2022a, SurveyVehicle2024}.
Drifting has emerged as a crucial driving technique in this field, allowing vehicles to achieve high-speed turning at sharp corners by intentionally inducing controlled oversteering and traction loss \cite{RealTimeDriftDriving2022}.
This driving technique offers promising solutions for stabilizing autonomous vehicles under extreme driving situations, e.g., such as sudden oversteering during high-speed obstacle avoidance or skidding on slippery roads, which can result in high sideslip angles or large yaw rates similar to drifting states. 
In recent years, the pursuit of high-performance driving experiences has led to a surge in research interest in autonomous drifting, which can be categorized into sustained drift, transient drift, and inertial drift. 
Researchers have discovered two saddle points outside the vehicle stability region through phase portrait analysis, namely drift equilibrium points (DEP), whose existence provides steady-state cornering conditions for drifting vehicles \cite{SteadystateDrifting2011a, EquilibriumAnalysis2010a, VehicleDrifting2022}. 
Sustained drift focuses on maintaining the vehicle states close to these unstable DEP to follow a circular path. 
Transient drift requires temporary drifting maneuvers to alter the driving direction rapidly for applications of drift parking \cite{AutonomousDrift2017a} or cornering \cite{DriftControl2018c}. 
While inertial drift requires transitioning through a series of transient states to navigate through consecutive turns without speed loss\cite{ConsecutiveInertia2023a}.

All drifting categories share the common objective of path tracking while maintaining controlled drift states for a long or short period.
Conventional path tracking control strategies, such as pure pursuit and the Stanley controller, compute the steering angle based on the vehicle's position \cite{ReviewPerformance2021a}, which are inadequate for drift vehicles due to their distinctive attributes of the opposite steering angle and yaw rate. 
Kapania et al. \cite{DesignFeedbackfeedforward2015} and Goh et al. \cite{SimultaneousStabilization2016} both designed a controller with drifting and path-tracking states as the controller references, however, the control commands for precise path tracking may cause the vehicle to exit drifting states \cite{CombinedFast2022a}.
Hierarchical control frameworks are utilized to separately deal with the inherent conflict control objectives between drifting and path tracking.
Chen et al. \cite{DynamicDrifting2023} employed a hierarchical control framework including a path tracking layer, vehicle motion control layer, and actuator regulating layer.
The path tracking layer utilizes a model predictive controller (MPC) to solve a finite horizon optimal control problem for the desired control variables.
Hu et al. \cite{NovelModel2024} constructed an optimization problem to determine the best future trajectory with minimized path tracking error.
These approaches necessitate the prediction of vehicle trajectory in forthcoming steps, namely prediction-based path tracking (PPT) control methods, which can significantly escalate the computational burden due to the successive prediction procedures.

Another drawback of existing research is that they rely heavily on precise system models to derive the DEP, which can be a huge challenge for drifting vehicles in two folds.
First, the significant nonlinearity of drifting states and tire force makes it difficult to create accurate analytical representations. 
To release the computation burden of the complex vehicle dynamics, Hu et al. \cite{CombinedFast2022a} used a Taylor expansion to linearize the original nonlinear vehicle model around an equilibrium point, establishing the error dynamic model.
Similarly, Shi et al. \cite{NonlinearModel2023} adopted Koopman operator theory and dynamic mode decomposition to obtain an approximately linear model.
Though such linear models can locally represent system dynamics, they still exhibit modeling errors compared to the actual system, which can potentially degrade system performance.
Second, vehicle parameters are hard to identify in real-time due to their sensitivity to environmental changes or tire conditions. 
The interaction of complex nonlinear drifting dynamics and misidentified physical parameters can aggravate the modeling difficulty, resulting in biased DEP to degrade controller performance.

Learning-based approaches have been involved in relaxing the stringent requirements in the modeling process. 
Joa et al. \cite{NewControl2020} mimicked the expert driving and Acosta et al. \cite{TeachingVehicle2018c} trained a neural network to provide the desired drifting states for the controller, eliminating the need for prior knowledge of DEP and tire parameters. 
Besides, deep reinforcement learning approaches \cite{HighSpeedAutonomous2020a} have also been utilized to learn control policies through interacting with the environment, which can perform stable drifting maneuvers even in the unseen map.  
These model-free learning methods exhibit excellent generalization ability to various road terrains but are computationally complex and data-hungry.
Besides, the training process detached from vehicle dynamics can be potentially dangerous, especially for high-speed operating vehicles.

\begin{figure}
	\centering
	\includegraphics[width=1\columnwidth]{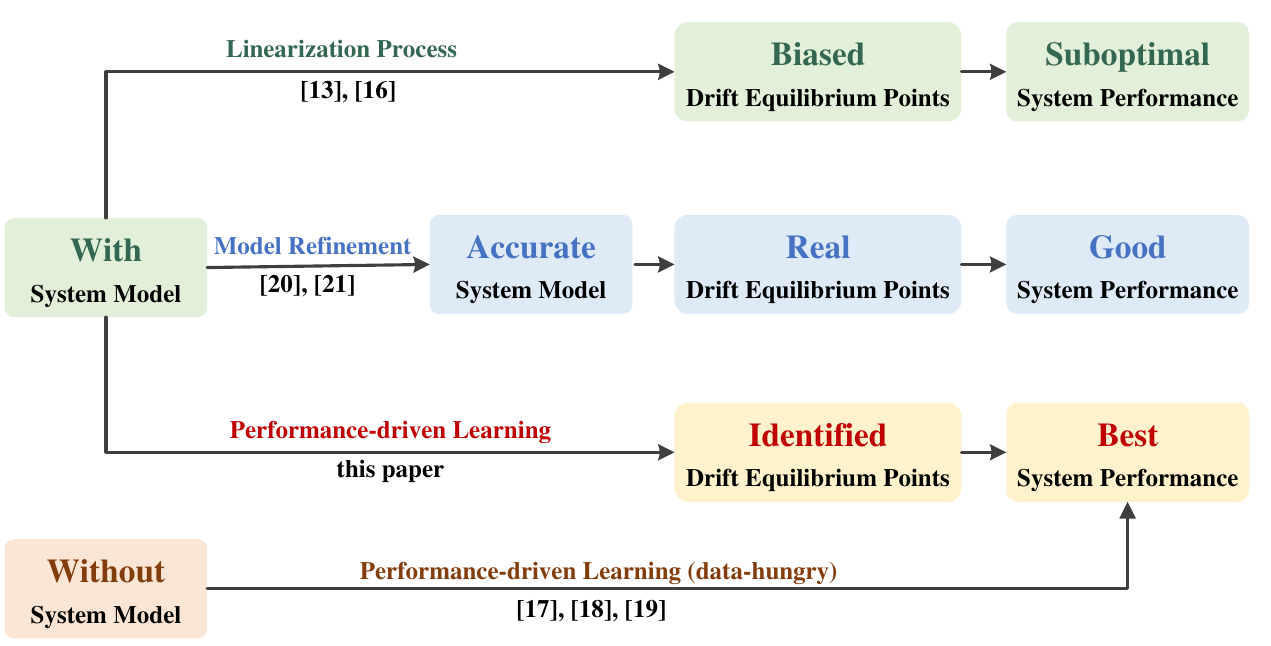}
	\caption{The principle of DEP identification.}
	\label{IEQ}
\end{figure}

Model-based learning approaches perform by refining the initially established system dynamics through real-world learning. 
Jia et al. \cite{NovelNonlinear2023} established a novel dynamic model for drift motion, which changed the equilibrium point analysis with saddle point properties to the nonlinear dynamic system.
Zhou et al. \cite{LearningBasedMPC2022a} learned the model bias with neural networks to enhance the precision of calculated drifting equilibrium.
However, according to the rationale of identification for control \cite{PerformanceOrientedModel2019}, high precision in system modeling does not necessarily assure the best output performance.
This suggests that the model refinement process can be sidestepped by directly identifying the DEP corresponding with the best driving performance, which is the motivation of our paper as illustrated in Figure~\ref{IEQ}.

Bayesian optimization (BO) is a model-based learning approach with high sample efficiency and global optimality, which has been extensively applied in hyper-parameter tuning for machine learning \cite{PracticalBayesian2012}, robotics \cite{BayesianOptimization2021}, and controller design \cite{ OptimalWeight2023}. 
The optimization process is guided by balancing exploration (sampling in unexplored regions) and exploitation (sampling in promising regions) for expensive-to-evaluate objective functions \cite{TakingHuman2016}, which can be a single goal like computing the minimal-time racing line \cite{ComputingRacing2020} or striking a harmonious balance among competing objectives \cite{MultiObjectiveOptimization2021, PredictivePath2023, PerformancebasedTrajectory2021}. 
BO can find great potential to bypass the sequential prediction process of PPT control methods by learning an adaptive path tracking (APT) control approach with minimized lateral error.
Besides, BO can also compensate for modeling errors and unknown disturbances by directly identifying the DEP, but not yet been utilized for drift vehicle control.

In this paper, BO serves as an upper-level supervisor to guide the lower-level drift control process in a performance-driven way: establish an APT control law and optimize the DEP. 
We choose MPC as the lower-level controller, which is a preferred vehicle control strategy owing to its superior capabilities to handle rapidly changing vehicle dynamics and manage constraints.
The main contributions are as follows:

\begin{enumerate}[label=\arabic*)]
    \item	The path tracking problem for drift vehicles is solved by executing smooth transitions among various drift states, and an APT control method is proposed to dynamically adjust the drift radius and steering angle to enhance the path tracking performance.

    \item   An adaptive learning-based model predictive control (ALMPC) strategy is proposed in this paper, where an upper-level BO supervisor is employed to compensate for modeling error by learning the APT control law and optimal DEP, then provides the learned parameters to instruct the lower-level MPC drift controller. 

    \item 	Simulation results indicate that the ALMPC strategy can effectively mitigate the drifting-tracking control conflict and compensate for modeling errors, outperforming traditional optimization-based approaches.
\end{enumerate}	

The remainder of this paper is organized as follows.
Section \uppercase\expandafter{\romannumeral2} introduces the preliminaries associated with drift vehicle control research to establish the groundwork for the proposed control strategy in the forthcoming section.
Section \uppercase\expandafter{\romannumeral3} elaborates the proposed {ALMPC} strategy, including the APT control law, DEP identification and the performance-driven system learning through BO.
The proposed ALMPC strategy is verified on the Matlab-Carsim platform in Section \uppercase\expandafter{\romannumeral4} and the conclusion is drawn in Section \uppercase\expandafter{\romannumeral5}.

\section{Preliminaries}
This section delves into the foundational aspects of drift vehicle control research, encompassing three key components: the MPC drift controller, PPT control strategy, and application of BO algorithms. 
The drift controller and PPT strategy aim to keep the vehicle close to the reference drifting states while minimizing the path tracking error to follow a predefined path. 
Besides, BO is applied as a pivotal tool for constructing a learning-based MPC framework to enhance drift control performance.

\subsection{Drift Controller Design}
The objective of the drift controller is to keep the vehicle states, such as the velocity, sideslip angle, and yaw rate to maintain at specific values, known as the DEP, to achieve optimal drift control.
These DEP will be derived from the vehicle dynamic model, which serves as the reference for the drift controller.

\subsubsection{Vehicle Dynamic Model}
\begin{figure}[t!]
	\centering
	\includegraphics[width=.8\columnwidth]{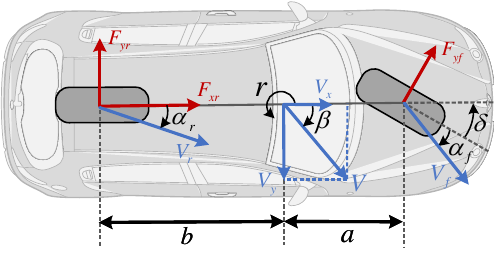}
	\caption{Single-track vehicle model for drifting.}
	\label{vehicle}
\end{figure}

Due to the symmetry of the vehicle, we adopt the single-track model to simplify the drift vehicle as a rigid body with mass $m$ and yaw inertia ${I_z}$, as depicted in Fig. \ref{vehicle}.
The single-track model is presented as follows:
\begin{align} 
	\dot{V} &= \frac{-F_{yf}\sin{(\delta-\beta)} + F_{yr}\sin{\beta} + F_{xr}\cos{\beta}}{m} \\
	\dot{\beta} &= \frac{F_{yf}\cos{(\delta-\beta)} + F_{yr}\cos{\beta} - F_{xr}\sin{\beta}}{mV} -r \\
	\dot{r} &= \frac{aF_{yf}\cos{\delta} - bF_{yr}}{I_z}
\end{align}

The model encompasses states of the vehicle absolute velocity $V$, sideslip angle $\beta$, and yaw rate $r$, all defined around the center of gravity (CoG).
$F_{yf}$ and $F_{yr}$ correspond to the lateral forces on the front and rear wheels, while $F_{xr}$ denotes the longitudinal forces on the rear tires.
$\delta$ represents the front wheel steering angle, $a$ and $b$ respectively signify the distances from the CoG to the front and rear wheel.

The complexity and nonlinearity of the vehicle model primarily stem from the tire force model, especially when the vehicle is in high-speed drifting states. 
The simplified Pacejka tire model \cite{TyreModelling1987} is typically used to  approximate the lateral tire forces $F_{yf}$ and $F_{yr}$, which closely matches the full Pacejka model but with significantly improved computational efficiency, as follows: 
\begin{align}
	F_{yi} &= -\mu F_{zi}\sin(C\arctan(B\alpha_{i}))  
\end{align}
where $\mu$ represents the coefficient of friction between the tire and ground, $F_{yi}$ and $ F_{zi}$ $(i = f, r)$ respectively represent the lateral tire forces and vertical loads on the front and rear tires, $B$ and $C$ represent the tire coefficients, $ \alpha_{i}$ $(i = f, r)$ represents the tire sideslip angles for the front and rear tires, which are formulated as: 
\begin{align} 
	\alpha_{f} &= \tan^{-1}(\frac{V\sin\beta + ar }{V\cos\beta}) - \delta\\
	\alpha_{r} &= \tan^{-1}(\frac{V\sin\beta - br }{V\cos\beta}) 
\end{align}

However, the longitudinal and lateral tire forces of a drift vehicle can not be represented independently by the Pacejka tire model due to their coupled relationship. 
For a rear-wheel-drive drift vehicle, the rear tire forces operate at the limits of friction, becoming saturated and pushing the vehicle into a fully sliding condition. Meanwhile, the front tire forces remain below saturation to maintain the vehicle's lateral stability and prevent it from spinning out.
To capture this high level of coupled tire force, the friction circle can be used to derive the rear tire force as follows \cite{AutomatedVehicle2019}:
\begin{align} 
	F_{yr} &= \sqrt{(\mu F_{zr})^2 - F_{xr}^2 }
\end{align}
This equation considers the limited friction to derive the maximum lateral force that the rear tire can generate given the longitudinal force.

\subsubsection{Drift Equilibrium Analysis}
Drift vehicle control hinges on maintaining the vehicle states around the unstable DEP, which can be obtained by setting all state derivatives of the vehicle model to zero as $\dot{V}$ = $\dot{\beta} $ = $\dot{r}$ = 0  \cite{ControllerFramework2014a}.
The initial steering angle is assumed to maintain a constant value at $\delta^{eq}$ to facilitate deep drift states. 
The desired drift radius $ R^{eq} $ can also be acquired from the proposed APT control method, establishing a relationship between the desired drift velocity $V^{eq} $ and yaw rate $r^{eq} $ as $R^{eq}  = V^{eq}  /r^{eq} $. 
Then, other desired DEP can be obtained by solving the state derivative equations.
The notations in this paper are summarized in Table \ref{notation}.
	
\begin{table}[width=1.0\linewidth,cols=4,pos=b]
\caption{Symbols and Descriptions \label{notation}}
\centering
\begin{tabular*}{\tblwidth}{l|l}
	\toprule
	\textbf{Acronym} & \textbf{Description}  \\
	\midrule
	\textbf{DEP} & Drift Equilibrium Points \\
	\textbf{PPT} & Prediction-based Path Tracking \\
	\textbf{APT} & Adaptive Path Tracking \\
	\textbf{ALMPC} & Adaptive Learning-based MPC \\
	\midrule
	\midrule
	Notation & Description  \\
	\midrule
	$ R^{eq} $ & Desired drift radius \\
	$\bm{x}_k = [V, \beta, r]^T$ & Vehicle states \\
	$\bm{u}_k = [\delta, F_{xr}]^T$ & Vehicle input variables \\
	$\bm{\hat\xi_k} = [\bm{x}_{k+1}, \bm{u}_k]^T $ & Drift states\\
	$\bm \xi^{eq}   =  [ V^{eq}  ,\beta^{eq}, r^{eq},  \delta^{eq},  F_{xr}^{eq}  ] $ & DEP \\
	$e_{la}$ &  Look-ahead error\\
	$x_{la}$ & Look-ahead distance \\
	$R^{eq} = w_r R_r   + w_e e_{la}$ & APT control law\\
	$\hat \delta^{eq} = \delta^{eq}  + ke_{la}$ & Feedback control law \\
	$\bm{\theta} = [ \delta^{eq},  w_r, w_e]$ & Parameters learned in BO \\
	$ \bm{\theta}^* =  [{\delta^{eq}}^*, {w_r}^*, {w_e}^*]$ & Optimal system parameters \\
	
	\bottomrule
\end{tabular*}
\end{table}
	
Nonetheless, it is noteworthy that the calculated DEP is prone to deviate from the real ones due to inaccurate system modeling.
Performance-driven approaches are adopted to adjust the calculated DEP, with more details in the following section.

\subsubsection{MPC formulation}
MPC is a preferred control strategy for autonomous vehicles, which employs a dynamic system model to anticipate future behavior within a finite time horizon, thereby determining the optimal control inputs.
In MPC, the control problem is formulated as an optimization problem to minimize a cost function subject to system dynamics and constraints.
To facilitate this, the vehicle model is linearized around the DEP as:
\begin{equation}
	\begin{aligned} 
		\bm{x}_{k+1} &= \bm{Ax}_k + \bm{Bu}_k + d \\
		d & = \bm{x}^{eq}   -  \bm{A}\bm{x}^{eq}   -  \bm{Bu}^{eq} \\	 
	\end{aligned} 
\end{equation}
where the vehicle states are denoted as $\bm{x}_k = [V, \beta, r]^T$, the vehicle input variables are denoted as $\bm{u}_k = [\delta, F_{xr}]^T$, and 
$d$	is a constant term that compensates for the deviation during the linearization process around the DEP.
This linearized MPC model can achieve good accuracy to provide a reliable basis for accurate short-term predictions.

Drift states are defined as $ \bm{\hat\xi_k} = [\bm{x}_{k+1}, \bm{u}_k]^T $ to minimize the cost function by constructing a quadratic programming problem as 
\begin{align} 
	\bm{\hat\xi}_{k+1} & = \bm{\hat{A}}\bm{\hat{\xi}}_k + \bm{\hat{B}}\Delta \bm{ u_k} + \bm{\hat {D}}  
\end{align}
$\Delta \bm{u}_k = \bm{u}_k - \bm{u}_{k-1}$ represents the input change between two consecutive time steps.
$\bm{A}, \bm{B} $ are the Jacobian matrices calculated at the DEP,  while $\bm{\hat {A}}, \bm{\hat {B}} $ and $\bm{\hat {D}}$ are represented as 
\begin{align*}
	\bm{\hat {A}} =\begin{bmatrix} \bm{A} & \bm{B} \\ \bm 0  & \bm I \end{bmatrix}, \
	\bm{\hat {B}} =\begin{bmatrix}   \bm B \\  \bm I \end{bmatrix}, \
	\bm{\hat {D}} =\begin{bmatrix}   d \\  \bm 0 \end{bmatrix}  
\end{align*}

Due to limitations of vehicular hardware, we have constraints on the inputs $\bm u \in \mathcal{U}: = \{\delta_{\text{min}} \le \delta \leq \delta_{\text{max}}, F_{\text{min}} \leq F_{xr} \leq F_{\text{max}} \}$.
Additionally, we have constraints on the input changes $\Delta \bm u \in  \mathcal{V}:  \{ |\Delta \delta| \le \Delta \delta_{\text{lim}}, |\Delta F_{xr}| \le \Delta F_{\text{lim}} \}$ to ensure safe and smooth driving.

In this paper, a hierarchical control structure is proposed to separate path tracking and drifting control, thereby the target of MPC is to maintain the vehicle states at the DEP $ \bm \xi^{eq}   =  [ V^{eq},\beta^{eq}, r^{eq},  \delta^{eq},  F_{xr}^{eq}  ] $.
The discrete-time optimization problem is formulated as 
\begin{align} 
	\text{min}\sum\limits_{k=1}^{N_p} ||\bm {\hat \xi}_k- \bm \xi^{eq}  ||^2_{\bm Q} + \sum\limits_{k=1}^{N_c} ||\Delta \bm u_k||^2_{\bm R} 
\end{align} 
\begin{align*} 
	\text{s.t.} \quad
	\bm {\hat\xi}_{k+1} &= \bm{\hat {A}\hat{\xi}}_k + \bm {\hat {B}}\Delta \bm u_k + \bm {\hat {D}} \\
	\bm u  &\in \mathcal{U}, \ \Delta \bm u \in \mathcal{V} 
\end{align*} 
In this formulation, $N_p$ and $N_c$ represent the selected predictive and control horizons, $\mathbf{Q} $ and $\mathbf{R} $ are positive definite weighting matrices.
The first part of the objective function is to keep the vehicle close to the reference drifting states, while the second part is to avoid abrupt changes in input to promise safety and driving comfort.

\subsection{Prediction-based Path Tracking Control}
\begin{figure}[!ht]
	\centering
	\includegraphics[width=.9\columnwidth]{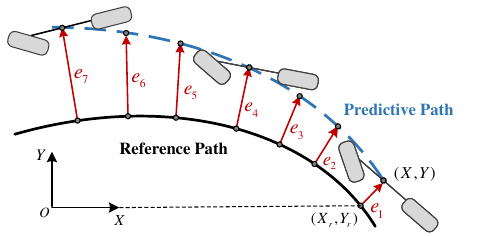}
	\caption{Prediction-based path tracking (PPT) control.}
	\label{PPT}
\end{figure}

Since different sustained drift states result in circular paths with varying radii, the vehicle can transition among various drift states to follow an arbitrary reference path \cite{HybridHierarchical2017}. 
PPT control approaches employ a circular path to fit the reference path in several forthcoming steps, then solve an optimization problem to find the optimal drift radius with minimum lateral errors in the predictive horizon \cite{ MPCbasedController2021a}, as shown in Figure \ref{PPT}.  

However, these approaches are highly dependent on accurately predicting the vehicle's position and require substantial computational resources, which can potentially impact real-time performance.
To address this problem, an APT path tracking strategy is proposed in this paper with more details in the following section.

\subsection{Bayesian Optimization}
Bayesian optimization is a data-efficiency strategy suitable for finding the global optimum of an expensive-to-evaluate function whose derivatives and convexity properties are unknown \cite{TutorialBayesian2010}.
The optimization problem is formalized as
\begin{align} 
	\bm{\theta^*} &= \arg\min_{\bm{\theta \in \Theta}} J(\bm{\theta})
\end{align}	
where the goal is to find the best set of parameters $\bm{\theta^*}$ that minimizes a cost function $J(\bm{\theta})$, with $\bm{\theta}$ belonging to a set of possible parameters $\Theta$.

Bayesian optimization consists of two essential ingredients: the probabilistic surrogate model and the acquisition function.
The probabilistic surrogate model provides posterior distribution and predictions over the black-box objective function $J(\bm{\theta})$.
Based on the probabilistic prediction, the acquisition function guides the optimization process by determining the next evaluated point.

Nonparametric Gaussian Process (GP) model \cite{GaussianProcesses2004} serves as a powerful tool to characterize BO objective function as
\begin{align}
	J(\bm{\theta}) \sim \mathcal{GP}(\mu(\bm{\theta}), {\sigma}^2(\bm \theta))
\end{align}
where $\mu(\theta)$ is its mean function and ${\sigma}^2(\bm \theta)=k(\bm{\theta}_i,\bm{\theta}_j)$ is the covariance function indicating the correlation between $J(\bm{\theta}_i)$ and $J(\bm{\theta}_j) $ , which is also known as the kernel function. 
The core of GP lies in the choice of kernel function since it determines the functional characteristics, such as smoothness and correlation between different points in the parameter space.
Common kernels include the Gaussian kernel, exponential kernel, Mat\'ern kernel, etc. 

Assume that we have an initial set $\mathcal{D}_n = \{\bm{\theta}_i, \hat J(\bm{\theta}_i)\}^n_{i=1}$ to build the GP model, where $\theta_i$ is the input parameters and $\hat J(\bm{\theta}_i) = J(\bm{\theta}_i) + \epsilon$ are their corresponding noisy observation of true function values $J(\bm{\theta}_i)$ with Gaussian-distributed noise $\epsilon \sim N(0,\sigma_w^2) $. 
Given a test point $\bm{\theta}^*$, GP make predictions of the function value $J(\bm{\theta}^*)$ by given its mean function $\bm{\mu}(\bm \theta^*)$ and covariance function $\bm{\sigma}^2(\bm \theta^*)$ as
\begin{align}
	\bm{\mu}(\bm \theta^*) &= \mu(\bm{\theta}) + \mathbf{k_n}(\mathbf{K_n} +\sigma_w^2\mathbf{I_n})^{-1}(\hat{J}_n - \mu(\bm{\theta})) \\
	\bm{\sigma}^2(\bm \theta^*) &= k(\bm{\theta}^*, \bm{\theta}^*) - \mathbf{k_n}(\mathbf{K_n} + \sigma_w^2\mathbf{I_n})^{-1}\mathbf{k_n^T}
\end{align}
where $\mathbf{K_n} $ is the covariance matrix with entries $[\mathbf{K_n}]_{(i,j)} = k(\bm{\theta}_i, \bm{\theta}_j)$ and
the vector $\mathbf{k_n} = [\mathbf{k_n}(\bm{\theta}^*, \bm{\theta}_i)]$ 
contains the covariances between the new input $\theta^*$ and the observed data points in $\mathcal{D}_n,  i,j \in \{1,...n\}$. 

Then, the acquisition function makes use of the predictive distribution to select the next evaluated point with a balance between exploring new regions and exploiting promising areas.
Commonly used acquisition functions include the expected improvement (EI), probability of improvement (PI), and upper confidence bound (UCB).

\begin{figure*}[!t]
	\centering
	\includegraphics[width=2.05\columnwidth]{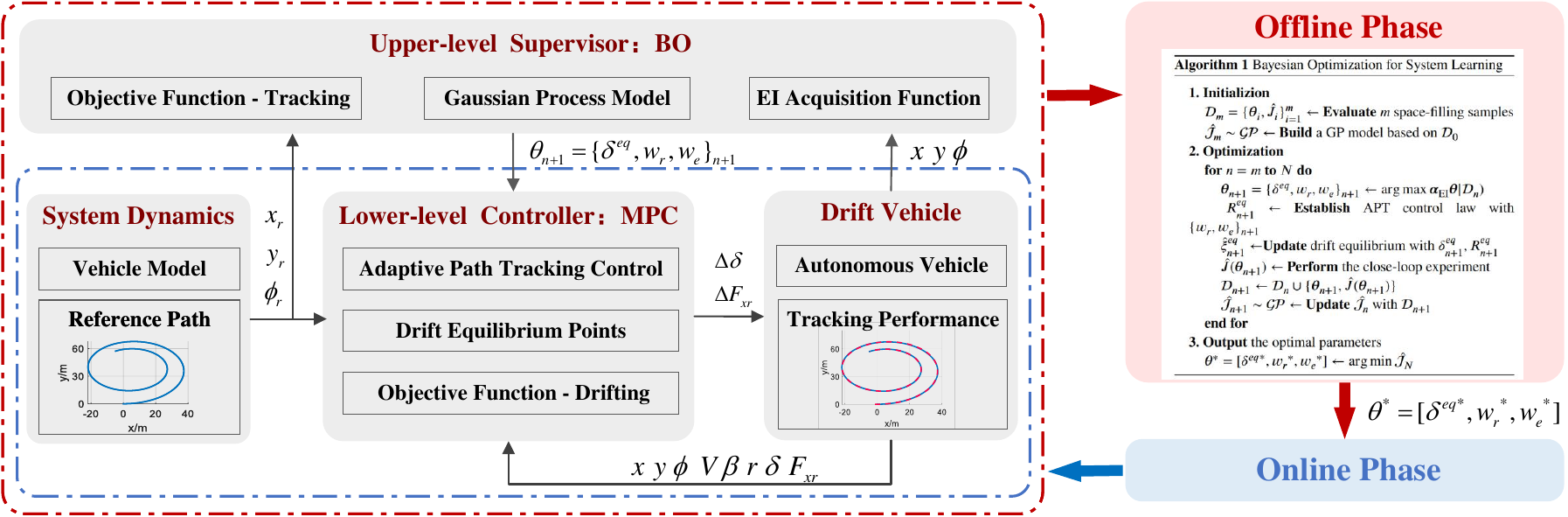}
	\caption{Hierarchical system architecture of the ALMPC strategy.}
	\label{Revise_BO}
\end{figure*}

\section{ The Proposed Adaptive Learning-based Model Predictive Control Strategy}

In this section, we will introduce the proposed ALMPC strategy, which is composed of an APT control law to minimize tracking error and DEP modification to enhance drift performance.
An upper-level BO supervisor is applied to learn the optimal APT control law and DEP in a performance-driven way, then provides these learned system parameters to instruct the lower-level MPC drift controller.
The hierarchical system framework is illustrated in Fig. \ref{Revise_BO}.

\subsection{ Adaptive Path Tracking Control}
To handle the path tracking error, an APT control law is proposed to dynamically adjust drift vehicle states to follow the reference path.
It can significantly alleviate the computation burden in each time step compared with the previously mentioned PPT control method. 
The proposed APT method conducts dynamic adjustments in two ways: direct adjustment of the drift radius and indirect adjustment of the steering angle.

\subsubsection{Drift Radius Calculation}
\begin{figure}[!t]
	\centering
	\includegraphics[width=.9\columnwidth]{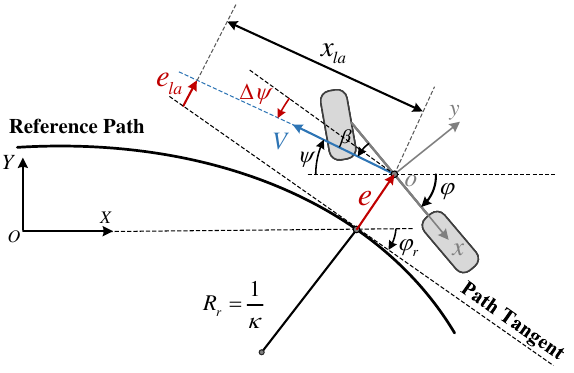}
	\caption{The proposed adaptive path tracking (APT) control.}
	\label{APT}
\end{figure}

Given the assumption that the reference path can be approximated as a series of circular arcs, the drifting radius at each time step technically should align with the radius of the corresponding arc.
Nevertheless, variations stemming from control limitations and environmental changes may cause the vehicle to deviate from the predefined path, emphasizing the necessity to dynamically adjust the drift radius.
We opt for the look-ahead error $e_{la}$ as the adjustment criterion, which is the predicted lateral deviation projected at a look-ahead distance $x_{la}$ ahead of the vehicle as
\begin{align} 
	e_{la} &= e + x_{la}\sin \Delta{\psi} 
\end{align} 
where $e = \sqrt{(X_r - X)^2 + (Y_r - Y)^2}$ denotes the tracking error from the vehicle's current position $(X, Y)$ to the closest point on the reference path $(X_r, Y_r)$, $\Delta \psi = \Delta \varphi + \beta$ denotes the course direction error, $\Delta \varphi = \varphi - \varphi_r $ denotes heading direction error between the vehicle's heading direction angle $\varphi$ and the reference angle $\varphi_r$, all in the global coordinate system as depicted in Fig. \ref{APT}.

The anticipated $e_{la}$ can consider both the lateral error and the course direction error, and only if simultaneously mitigates these two errors can the vehicle continually adhere to the reference path.
A positive look-ahead error $e_{la}>0$ signifies the future position of the vehicle will fall inside the reference path, necessitating a larger drift radius $R_r < R^{eq}  $ at the subsequent time step.
Conversely, a negative look-ahead error $e_{la}<0$ signifies a smaller drift radius.
Therefore, the desired drift radius $ R^{eq} $ for the upcoming step can be designed as: 
\begin{align} 
	R^{eq}    &= w_r R_r   + w_e e_{la} 
\end{align}	
where $R_r  $ is the radius of the approximated circular arc, which can be calculated from the current curvature of the reference path $\kappa$ as $ R_r  = 1/\kappa $.
By carefully tuning the weights $w_r$ and $ w_e $, the vehicle can adaptively adjust its drift radius according to current tracking performance.

\subsubsection{Steering Angle Feedback}
The adaptive adjustment on steering angle draws inspiration from the driving skills of professional racers: attaining a larger drift radius requires an increased steering angle to reach the equilibrium point $\delta^{eq}$. 
In the proposed ALMPC, a feedback law will dynamically adjust $\delta^{eq}$ at each time step to achieve the real-time desired steering angle $\hat \delta^{eq}$.

The look-ahead error $e_{la}$ serves as the adjustment criterion to design the feedback control law instead of the lateral error $e$ as:
\begin{align}
	 \hat \delta^{eq} = \delta^{eq}  + ke_{la}
\end{align}
where $k$ denotes the proportional gain and $ \hat \delta^{eq} $ denotes the adjusted steering angle equilibrium point for controller. 
The feedback control law can capture potential path deviation and allow for preemptive adjustment on controller inputs, which enhances control performance and overall stability.

\subsection {Drift Equilibrium Points Identification }

The effectiveness of MPC drift control is primarily dependent on precise system models to calculate the DEP.
However, this modeling process can be a formidable challenge due to the fluctuating road conditions and significant nonlinear drifting dynamics.
To tackle this problem, DEP identification is involved to compensate for the modeling error and undetermined disturbances.
Such practice can skip the laborious process of system model refinement and directly provide the modified reference to affect the MPC drift controller performance.

The idea of identification for control illustrates that the optimal DEP may not necessarily align with the real-world dynamics, but provide the best output performance.
In line with this perspective, the modification step endeavors to identify 'DEP' according to the vehicle's real-time performance.
These identified points should reflect the best vehicle performance, but may not necessarily correspond to the real DEP derived from the accurate system model.

Since the DEP  $\xi^{eq}$  = [ $V^{eq}$, $\beta^{eq}$, $r^{eq}$, $\hat \delta^{eq}$, $ F^{eq}$ ] are calculated with an initial assumption of $\delta^{eq} $, whose value has the greatest impact on determining other points.
Therefore, we refer to  $ \delta^{eq}  \in \mathbb{R}$ as the only identified drift equilibrium point, and others still be calculated from the system model.
This can partly leverage the rationality in system models to guide the identification process and release computation burden, since the existence of modeling error will mislead the calculated DEP, but remain close to the optimal ones.
If we simultaneously identify several equilibrium points detaching the system model, the estimation process can be aggressive and may even move in the opposite direction.

In this paper, we employ BO to explore the relationship between DEP and vehicle performance and the specific identification process will be illustrated in the following section.

\subsection {Performance-driven System Learning through Bayesian Optimization}

BO is a global optimization technique for expensive-to-evaluate black-box functions, which is suitable for learning the ambiguous relationship between DEP and vehicle performance.
Considering the limited prior knowledge of the proposed APT control law, the weights $w_r$ and $ w_e$ are also two critical parameters remaining to be learned via BO.
Therefore, the set of parameters to be learned through BO is composed of two parts correlated with the DEP and the proposed APT control law, denoted as $\bm{\theta} = [ \delta^{eq},  w_r, w_e]  \in \mathbb{R}^3$.
BO works as an upper-level supervisor to discover these optimal parameters through real-world experiments, and then provides the learned parameters to instruct the lower-level MPC.

Since maintaining drifting states has been assigned as the main goal of the lower-level MPC, the learning process of BO is guided by evaluating the path tracking performance through the lateral error $ e_k$ and course direction error $\Delta \psi_k$.
The optimization problem is constructed as
\begin{align} 
	J(\bm{\theta}) &= \text{log}  \big[ \frac{1}{N_k} \sum\limits_{k=1}^{N_k} (| e_k| + \lambda|\Delta \psi_k| ) + B(e_k) + I(e_k) \big]
\end{align} 
where
\begin{align} 
	B(e_k) &= \text{log} \big[\frac{1}{N_k} \sum\limits_{k=1}^{N_k} [10(e_k - e_{max})^+] \big] \\
	I(e_k) &= \frac{1}{N_{k-1}} \sum\limits_{k=1}^{N_{k-1}} (e_{k+1} - e_{k}) 
\end{align}
$B(e_k)$ is the soft barrier function restricting the lateral deviation to ensure safety and accelerate the optimization process. 
$I(e_k)$ is the increment of vehicle lateral deviation between two consecutive time steps, considering the motion sickness and potential danger caused by rapidly changing position. 
The term $[a]^+ = \max[a,0]$ indicates the positive part exceeding the desired maximum lateral error $e_{max}$.
$N_k$ represents the control period along the path and $\lambda$ is a constant parameter to govern the trade-off between these two indices.

For the GP surrogate model, we assume the zero mean function $\mu(\bm{\theta}) = 0$  and employ the Mat\`{e}rn kernel with parameter $\nu = 5/2 $, defined as
\begin{align}
	k(\bm{\theta}_i, \bm{\theta}_j) &= \sigma_{\eta}^2(1+\sqrt{5}\rho + \frac{5}{3}\rho^2) \text{exp}(-\sqrt{5}\rho)
\end{align}
where $\sigma_{\eta}^2$ is the prior covariance function,
$\rho = \frac{||\bm{\theta}_i - \bm{\theta}_j||}{l}$  is the scaled distance between two $\bm{\theta}_i$ and $\bm{\theta}_j$, and 
$l$ is the length scale parameter that determines the influencing range of the kernel.

\begin{algorithm}[!b]
\caption{Bayesian Optimization for System Learning }
\label{alg:alg1}
\setstretch{1.1}
\begin{algorithmic}
    \STATE 
    \STATE{\textbf{1.\ Initializion}  }\\ 
    
    \hspace{0.3cm} $\mathcal{D}_m = \{\bm{\theta}_i, \hat J_i\}_{i=1}^m \gets$ \textbf{Evaluate} $m$ space-filling samples\\
    
    \hspace{0.3cm} $	\hat{ \mathcal{J}_m} \sim \mathcal{GP} \gets$ \textbf{Build} a GP model based on $\mathcal{D}_0$\\
    
    \STATE{\textbf{2.\ Optimization} } 
    \STATE \hspace{0.3cm} {\textbf{for}} $n = m \ \textbf{to} \  N$ \textbf{do}\\
    
    \hspace{0.7cm} $\bm{\theta}_{n+1} = \{\delta^{eq},  w_r, w_e\}_{n+1}  \gets \arg\max \bm{\alpha}_{\text{EI}} \bm{\theta}|\mathcal{\bm{D}}_n)$ \\
    
    \hspace{0.6cm} $ R^{eq}_{n+1} $ $\gets$ \textbf{Establish} APT control law with $ \{ w_r, w_e\}_{n+1}$ \\
    
    \hspace{0.7cm}  ${\hat \xi}^{eq}  _{n+1} \gets $\textbf{Update} drift equilibrium with $\mathbf {\delta}^{eq}_{n+1}, R^{eq}_{n+1}$ \\
    
    \hspace{0.7cm} $\hat J(\bm{\theta}_{n+1}) \gets $ \textbf{Perform} the close-loop experiment \\
    
    \hspace{0.7cm} $\mathcal{D}_{n+1} \gets \mathcal{D}_{n} \cup \{\bm{\theta}_{n+1}, \hat J(\bm{\theta}_{n+1})\}$\\ 
    
    \hspace{0.7cm} $	\hat{ \mathcal{J}}_{n+1} \sim \mathcal{GP} \gets$ \textbf{Update} $\hat{ \mathcal{J}}_n$ with $\mathcal{D}_{n+1}$\\				
    
    \STATE \hspace{0.3cm} {\textbf{end for}}
    
    \STATE{\textbf{3.\ Output} the optimal parameters} 	 
    
    \STATE\hspace{0.3cm} $ \bm{\theta}^* =  [{\delta^{eq}}^*, {w_r}^*, {w_e}^*] \gets  \arg\min \hat{\mathcal{J}}_{N}$ \\		
		
\end{algorithmic}
\label{alg1}
\end{algorithm}

For the drifting MPC problem, we adopt the EI acquisition function defined as follows
\begin{equation}
	\begin{aligned} 
		\alpha_{\text{EI}}(\bm{\theta}|\mathcal{\bm{D}}_n) &= 
		\mathbb{E}\big[ [J(\bm{\theta})- J_n^* ]^+ \big]
	\end{aligned}
\end{equation}
where $\mathbb{E}[ \cdot]$ represents the expectation operator, $[ a]^+ = \max[a,0]$ indicates the positive part, and $J_n^* $ is the best observed value in historical samples $\mathcal{D}_n$. 
Furthermore, $\alpha_{\text{EI}}(\bm{\theta}|\mathcal{\bm{D}}_n)$ can be evaluated analytically using integration by parts \cite{EfficientGlobal1998}, described as
\begin{equation}
\begin{aligned} 
    \alpha_{\text{EI}} &= 
    \begin{cases}
        (\mu(\bm{\theta})-J_n^*)) \bm{\Phi}(Z)+ \sigma(\bm{\theta})\phi(Z) & \text{if } \sigma(\bm{\theta}) > 0\\
        0 & \text{if }  \sigma(\bm{\theta}) = 0
    \end{cases} \\
    Z & = \frac{\mu(\bm{\theta})- J_n^*}{\sigma(\bm{\theta})}  
\end{aligned}
\end{equation}
Then, the next evaluated set of parameters can be obtained through 
\begin{align}
	\bm{\theta}_{n+1} = \arg\max \alpha_{\text{EI}}(\bm{\theta}|\mathcal{\bm{D}}_n)
\end{align}
where $\phi(Z)$ and  $\bm{\Phi}(Z)$ denote the probability density function and the cumulative density function of the standard normal distribution, respectively.
The system learning process of BO is outlined in Algorithm \ref{alg:alg1}.

\begin{table}[width=1.0\linewidth,cols=4,pos=b!]
\caption{Vehicle and Control parameters in Simulation \label{parameters}}
\centering
\begin{tabular*}{\tblwidth}{cc|cc}
    \toprule
    Parameters  & Value & Parameters  & Value  \\
    \midrule

$m$ & \SI{1830}{\kg} & $\mathbf{Q}$ & \SI{}{\diag}(\SI{10}{},\SI{1}{},\SI{10}{},\SI{1}{},\SI{1}{})\\
$I_z$ & \SI{3234}{\unit{kg.m^2}} & $\mathbf{R}$ & \SI{}{\diag}(\SI{1}{},\SI{1}{}) \\
$a$ &  \SI{1.40}{\metre} & $N_p$ & \SI{20}{}  \\
$b$ & \SI{1.65}{\metre} & $N_c$ & \SI{19}{}   \\
$B_{f/r}$ & \SI{8.321}{} &  $T$  & \SI{18.4}{\second}\\
$C_{f/r}$ & \SI{1.626}{}  & $\Delta T $ & \SI{0.1}{\second}\\
$ \delta_{\text{min}}$ & \SI{-1}{\radian} &  $N_k$ & \SI{184}{} \\
$ \delta_{\text{max}}$ & \SI{1}{\radian} & $\kappa$& $\SI{1}{}/\SI{40}{}$\\
$ F_{\text{min}}$ & \SI{0}{\newton}& $\kappa'$ & $\SI{1}{}/\SI{12000}{}$ \\
$ F_{\text{max}}$ & \SI{9000}{\newton}  & $\theta_0$ & \SI{0}{} \\
$ \Delta\delta_{\text{lim}}$ & \SI{0.15}{\radian} &$(x_0, y_0)$ & $ (0, 0) $ \\
$ \Delta F_{\text{lim}}$ & 1000 \text{N} & $x_{la}$ & \SI{12}{\metre}  \\
$\delta_0$ &  \SI{-0.52}{\unit{\radian \per \second}} & $k$ & \SI{0.25}{}\\
$\mu $&  \SI{1}{} & $\mu_m $ &  \SI{0.9}{} \\
          
    \bottomrule
\end{tabular*}
\end{table}

\section{Simulation}
In this section, the proposed algorithm is verified in the Matlab-Carsim platform, which leverages Matlab's analytical system design tools alongside Carsim's realistic vehicle dynamics simulation.

\begin{figure}[!t]
\centering
\subfloat[ Clothoid-based path tracking. ]{\includegraphics[width=2.6in]{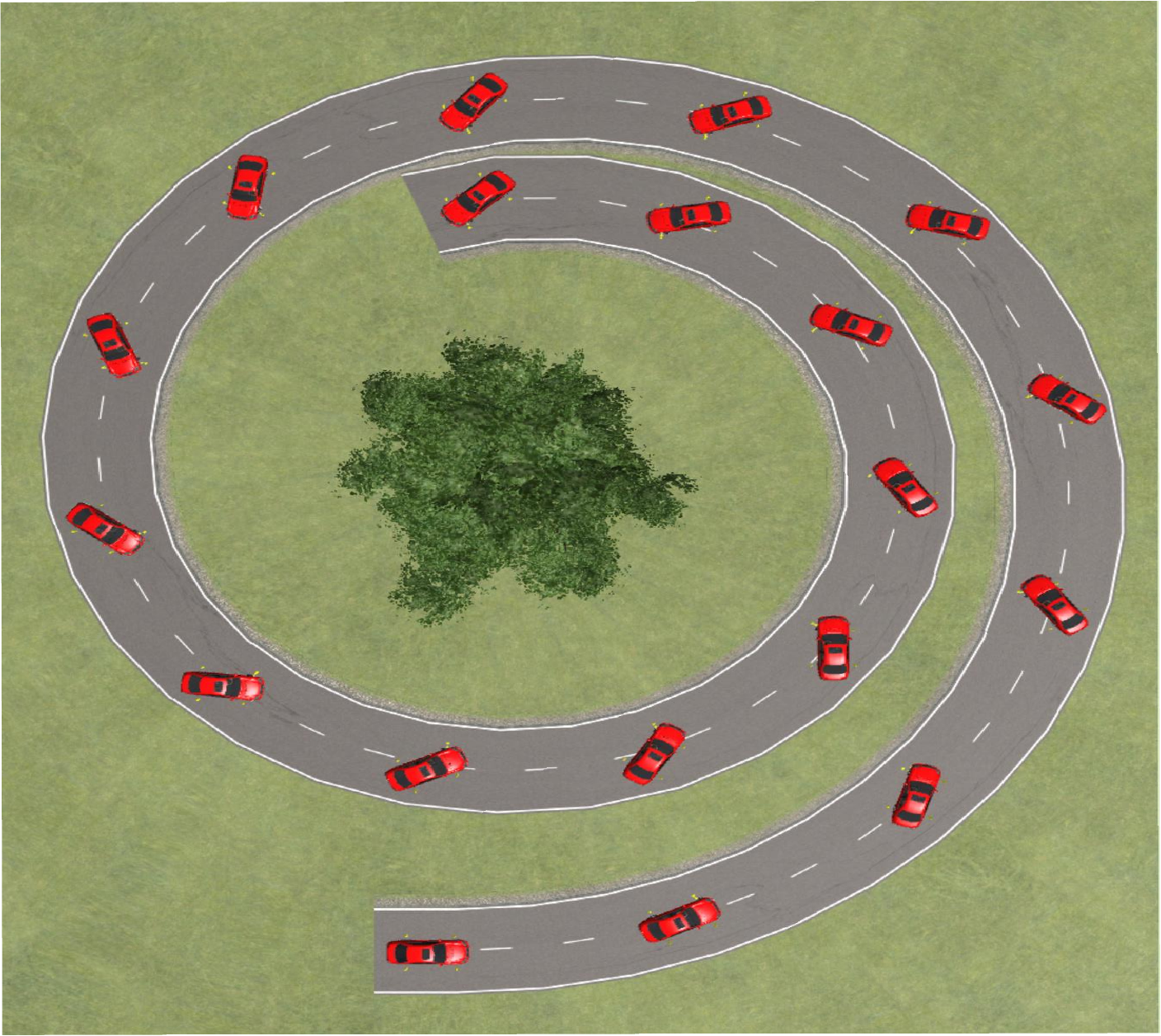}%
    \label{clothoid}}
\hfil
\subfloat[Drift Vehicle]{\includegraphics[width=2.6in]{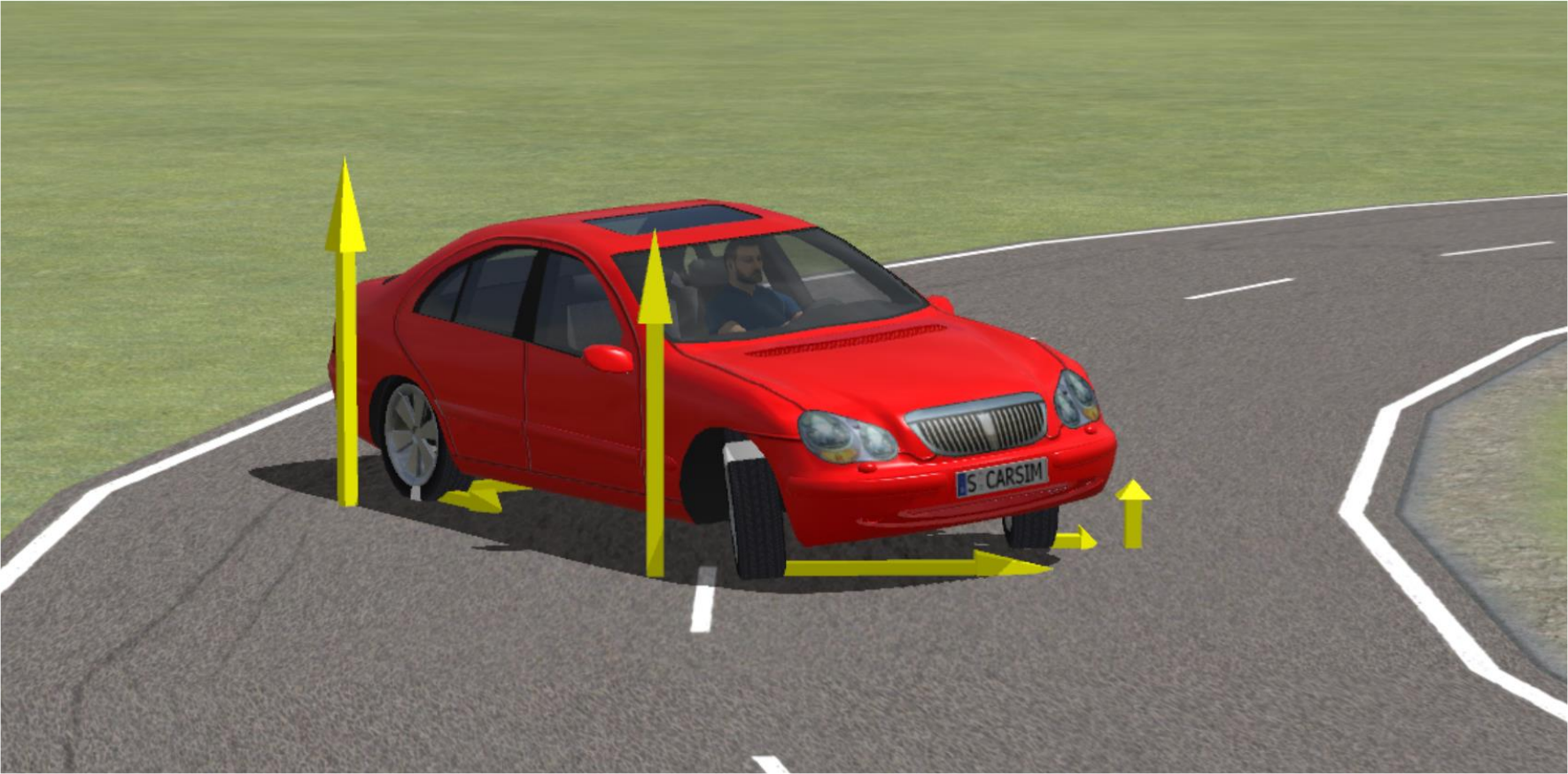}%
    \label{drift_vehicle}}		

\caption{ Clothoid-based path tracking in Matlab-Carsim platform.} 
\label{carsim}
\end{figure}

\begin{table*}[width=2.0\linewidth,cols=3,pos=t!]
\caption{Comparison of different simulations  \label{compare}}
\centering
\begin{threeparttable}
\begin{tabular*}{\tblwidth}{@{\extracolsep{\fill}}l|ll}
    \toprule
    Simulation   &  APT Learning - $w_r$ $w_e$ & DEP$^{1}$ Learning - $\delta^{eq}$\\
    \midrule
    MPC-PPT (baseline)& \textcolor{red}{\usym{2715}}$^{2}$ - PPT control& \textcolor{red}{\usym{2715}} - DEP derived from system model \\
    MPC-APT (proposed APT learning) & \textcolor[RGB]{0,160,0}{\usym{1F5F8}}$^{3}$ - learned APT control law & \textcolor{red}{\usym{2715}} - DEP derived from system model\\
    MPC-DEP (proposed DEP learning) & \textcolor{red}{\usym{2715}} \  - PPT control & \textcolor[RGB]{0,160,0}{\usym{1F5F8}} - learned DEP \\
    ALMPC (proposed APT \& DEP learning) & \textcolor[RGB]{0,160,0}{\usym{1F5F8}} \ - learned APT control law  & \textcolor[RGB]{0,160,0}{\usym{1F5F8}} - learned DEP \\
    \bottomrule
\end{tabular*}
    \begin{tablenotes}
        \footnotesize
        \item[1] DEP is short for DEP.
        \item[2] \textcolor{red}{\quad \usym{2715}} denotes the simulation operates without corresponding learning process.
        \item[3] \textcolor[RGB]{0,160,0}{\quad \usym{1F5F8}} denotes the simulation operates with learned parameters.  
        
    \end{tablenotes}
\end{threeparttable}
\end{table*}

\begin{table*}[width=2.05\linewidth,cols=14, pos=!t]
\small
\caption{Simulation Results for Case 1: Precise Vehicle Parameter \label{result1}}
\centering
\begin{threeparttable}
    \begin{tabular*}{\tblwidth}{@{\extracolsep{\fill}}c p{0.01cm} ccc p{0.01cm}cc p{0.01cm} ccccc}
        
        \toprule
        \multirow{3}{*}{ Simulation}  & &\multicolumn{3}{c}{System Parameters} & &\multicolumn{2}{c} {RMSE-Tracking $\downarrow$ } & &\multicolumn{5}{c}{RMSE-Drifting $\downarrow$}  \\
        \cline{3-5} \cline{7-8} \cline{10-14} 
        
        & &   $\delta^{eq}$  &    $w_r$ &   $w_e$ & &    $e$ &   $\Delta \psi$ & &   $V$  &   $\beta$ &   $r$ &   $\delta$ &   $F_{xr}$  \\
        
        & &   (rad) &     &    & &    (m) &   (rad) & &   (m/s) &   (rad) &   (rad/s) &   (rad) &   (N) \\
        
        \midrule
        
        MPC-PPT &  & $-0.520$ &  $-^1$ &  $-$ & &  $0.826$ &  $0.026$ & &  $0.595$ &  $0.208$&  $0.208$ &  $0.226$ &  $489.288$ \\
        
        MPC-APT &  & $-0.520$ &  $\mathbf{1.502}^2$ &  $\mathbf{0.890}$ & &  $0.567$ &  $0.024$ & &  $0.307$ &  $0.174$& \textcolor[RGB]{205,24,24}{ {$\mathbf{0.135}$}} &  $0.210$ &  $493.842$  \\

            MPC-DEP &  & $\mathbf{-0.461}$ &  $ - $ &  $-$ & &  $0.648$ &  $0.018$ & &  $0.537$ &  $0.208$& $0.162$ &  $0.253$ &  \textcolor[RGB]{205,24,24}{ {$\mathbf{489.137}$}}  \\
        
        ALMPC &  & $\mathbf{-0.482}$ &  $\mathbf{1.026}$ &  $\mathbf{0.945}$ & & \textcolor[RGB]{205,24,24}{ {$\mathbf{0.208}$}}$^3$ &  \textcolor[RGB]{205,24,24}{ {$\mathbf{0.015}$}} & & \textcolor[RGB]{205,24,24}{ {$\mathbf{0.264}$}} & \textcolor[RGB]{205,24,24}{ {$\mathbf{0.161}$}} &  $0.147$ & \textcolor[RGB]{205,24,24}{ {$\mathbf{0.198}$}}&   ${492.356}$  \\

        \bottomrule
    \end{tabular*}
    \begin{tablenotes}
        \footnotesize
        \item[1] '$-$' denotes that the parameter is not covered in the simulation.
        \item[2] Parameters learned through BO are highlighted in \textbf{bold}.
        \item[3] The best RMSE of each vehicle state is highlighted in \textcolor[RGB]{205,24,24}{\textbf{red bold}}.
    \end{tablenotes}
\end{threeparttable}
\end{table*}

\subsection{Simulation Setup}
The clothoid curve is a prevalent choice to fit the tracking path for autonomous vehicles, which has a gradually changing curvature to ensure a stable control process and minimized tracking deviation.
In this paper, we employed the clothoid-based path for stable drifting maneuvers, whose general parametric form can be expressed as follows:
\begin{align} 
	x &= x_0 + \int_0^{s} \cos(\frac{1}{2} \kappa' \tau^2 + \kappa\tau + \theta_0 )d\tau \\
	y &= y_0 + \int_0^{s} \sin(\frac{1}{2} \kappa' \tau^2 + \kappa\tau + \theta_0 )d\tau
\end{align} 
where $\kappa$ represents the curvature at the initial point of the path, $\kappa'$ represents the curvature changing rate, $s$ represents the arc length of the curve, $(x_0, y_0)$ represents the starting point of the vehicle, $\theta_0$ represents the initial angle, and $(x, y)$ represents the current position of the vehicle.
Fig. \ref{clothoid} illustrates the clothoid-based path tracking in the Matlab-Carsim platform, while the opposite orientations of the front wheel and the curved path in Fig. \ref {drift_vehicle} indicate that the vehicle is drifting.

In this paper, the objective of BO is evaluated over the whole path in every iteration for $T = 18.4 $s and the sampling time is set as $\Delta T = 0.1$ s, namely $N_k = 184$. 
We set $N_p = 20$ to provide a 2-second prediction horizon, which can balance stable vehicle performance with minimized computational burden.

According to the heuristics in \cite{rossetter2004potential}, the steering angle feedback in the APT control law employs $x_{la} = 12$ m and $k=0.25$ to strike a balance between a proactive response and an overreaction to minor disturbances. 
The initial steering angle is set as $\delta_0 = -0.52 $ rad ($-30^\circ$).
Other vehicle and controller parameters used in the simulation can be found in Table \ref{parameters}. 

Based on prior knowledge, the system parameters to be learned in the BO process are constrained to real-value in the interval $\delta^{eq} \in [-0.7, 0.4]$, $w_r \in [0, 2]$, and $w_e \in [-5, 5]$.
These constraints help guide the learning process towards practical and meaningful solutions.

Three comparisons are carried out to prove the effectiveness of ALMPC in both ways of drifting and tracking performance: MPC-PPT, MPC-APT, and MPC-DEP.
The comparison of these simulations is illustrated in Table \ref{compare}.
 
In MPC-APT, MPC-DEP, and ALMPC simulations, 20 random sets of learning parameters are generated to initialize the BO algorithm, and the optimization process is performed for 320 iterations. 

\subsection{Simulation Results}
To further validate the effectiveness of compensating modeling error, two case studies are carried out: the first case study employs precise vehicle parameters, while the second case intentionally uses misidentified vehicle parameters.

\subsubsection{Case 1: Precise Vehicle Parameter}
\begin{figure}[!b]
\centering
\subfloat[Lateral error]{\includegraphics[width=1\linewidth]{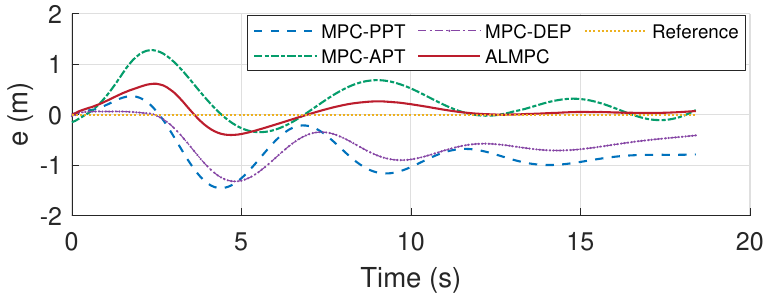}%
    \label{case2_e}}
\hfil
\subfloat[Course error]{\includegraphics[width=1\linewidth]{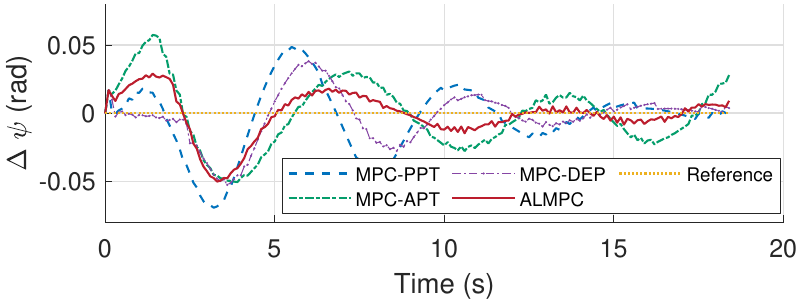}%
    \label{case2_zeta}}
\caption{ Simulations results of the path tracking performance in case 1: the proposed ALMPC significantly reduces tracking error compared to the baseline MPC-PPT owing to the incorporation of both APT and DEP learning. While simulations of MPC-APT and MPC-DEP incorporating partial learning steps can only achieve minor improvements. } 
\label{case2_tracking}
\end{figure}

\begin{figure}[!b]
\centering
\subfloat[Velocity]{\includegraphics[width=1\linewidth]{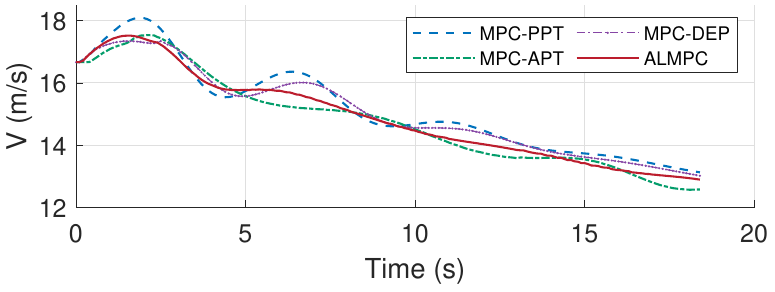}%
    \label{case2_v}}
\hfil
\subfloat[Sideslip angle]{\includegraphics[width=1\linewidth]{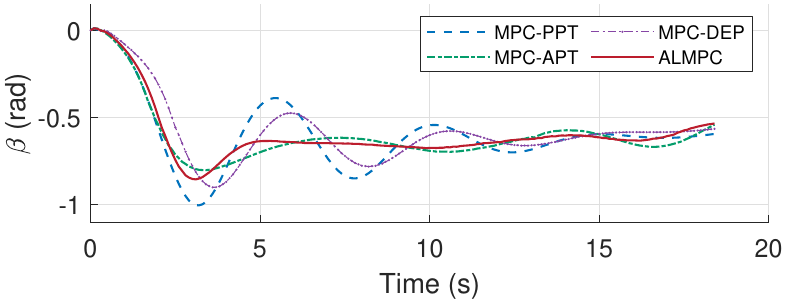}%
    \label{case2_beta}}
\hfil
\subfloat[Yaw rate]{\includegraphics[width=1\linewidth]{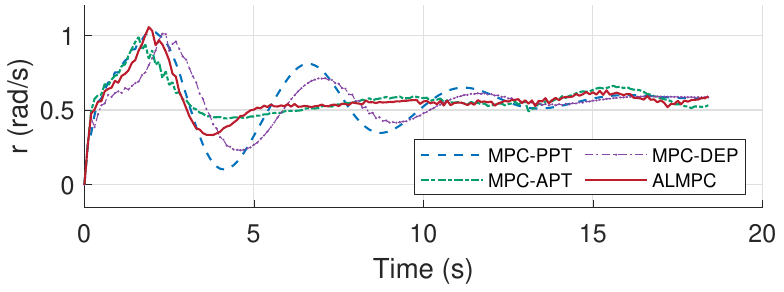}%
    \label{case2_r}}
\hfil
\centering
\subfloat[Steering angle]{\includegraphics[width=1\linewidth]{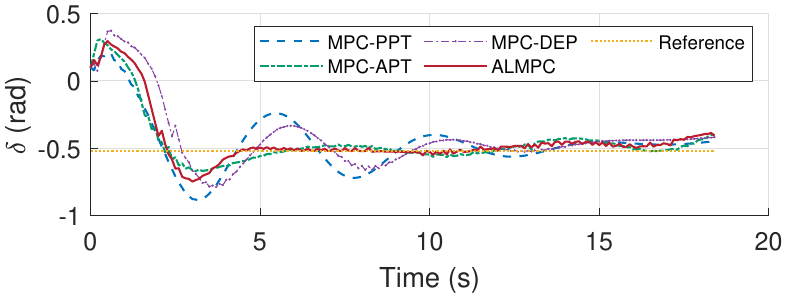}%
    \label{case2_delta}}
\hfil
\subfloat[Rear longitudinal force]{\includegraphics[width=1\linewidth]{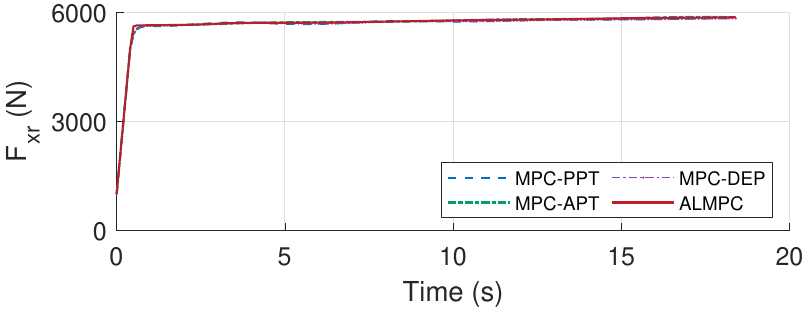}%
    \label{case2_F}}
\caption{Simulations results of drift vehicle states in case 1: the proposed ALMPC can achieve the best control performance with stable drift states than other simulations. } 
\label{case2_drifting_X}
\end{figure}

\begin{figure}[!t]
	\centering
	\includegraphics[width=0.9\linewidth]{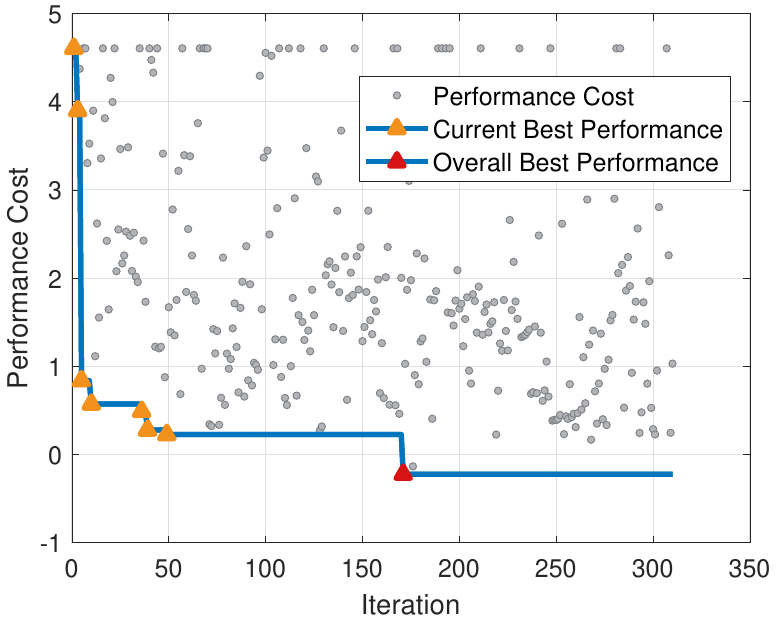}
	\caption{The path tracking performance cost during BO process in case 1: the lower path tracking performance cost points to the better path tracking performance and the tracking performance is gradually enhanced during the learning process and the optimal set of parameters corresponding to the best path tracking results is learned in iteration 171.  }
	\label{case2_BO_iterations}
\end{figure}

The system parameters and performance of three simulations are demonstrated in Table \ref{result1}, with the optimized parameters highlighted in bold and the best root mean square error (RMSE) for each vehicle state in red bold.
The RMSEs of $e$ and $\Delta \psi$ demonstrate the tracking performance to follow the given clothoid-based path.
Additionally, the RMSEs of $V$, $\beta$, $r$, $\delta$, and $F_{xr}$ illustrate the discrepancy between the vehicle's actual states and the planned drift states at each given time step, which effectively evaluates the drifting performance.

The path-tracking performance of the vehicle is shown in Fig. \ref{case2_tracking} and the drift vehicle states are shown in Fig. \ref{case2_drifting_X}.
Since each simulation features a unique planned series of drift states along the path, we only showcase the common reference used in all three simulations, which is $e=0$, $\Delta\psi=0$, and $\delta = -0.52$ rad.
The evaluation of other states relative to their reference can be found by referring to their respective RMSEs listed in Table~ \ref{result1}.

\begin{table*}[width=2.05\linewidth,cols=14, pos=t!]
\small
\caption{Simulation Results for Case 2: Misidentified Vehicle Parameter  \label{result2}}
\centering
\begin{threeparttable}
    \begin{tabular*}{\tblwidth}{@{\extracolsep{\fill}}c p{0.01cm} ccc p{0.01cm}cc p{0.01cm} ccccc}
        
        \toprule
        \multirow{3}{*}{ Simulation}  & &\multicolumn{3}{c}{System Parameters} & &\multicolumn{2}{c} {RMSE-Tracking $\downarrow$ } & &\multicolumn{5}{c}{RMSE-Drifting $\downarrow$}  \\
        \cline{3-5} \cline{7-8} \cline{10-14} 
        
        & &   $\delta^{eq}$  &    $w_r$ &   $w_e$ & &    $e$ &   $\Delta \psi$ & &   $V$  &   $\beta$ &   $r$ &   $\delta$ &   $F_{xr}$  \\
        
        & &   (rad) &     &    & &    (m) &   (rad) & &   (m/s) &   (rad) &   (rad/s) &   (rad) &   (N) \\
        
        \midrule
        
        MPC-PPT &  & $-0.520$ &  $-^1$ &  $-$ & &  $0.970$ &  $0.044$ & &  $0.751$ &  $0.277$&  $0.283$ &  $0.465$ & $410.755$  \\
        
        MPC-APT &  & $-0.520$ &  $\mathbf{1.527}^2$ &  $\mathbf{1.025}$ & &  $0.650$ &  $0.036$ & &  $0.399$ & $0.210$ & $0.128$ &  \textcolor[RGB]{205,24,24}{ {$\mathbf{0.208}$}} &  $410.776$  \\
        
        MPC-DEP &  & $\mathbf{-0.461}$ &  $ - $ &  $-$ & &  $0.794$ &  $0.033$ & &  $0.708$ &  $0.235$& $0.224$ &  $0.313$ &  $410.757$  \\
        
        ALMPC &  & $\mathbf{-0.471}$ &  $\mathbf{1.903}$ &  $\mathbf{1.032}$ & & \textcolor[RGB]{205,24,24}{ {$\mathbf{0.122}$}}$^3$ &  \textcolor[RGB]{205,24,24}{ {$\mathbf{0.010}$}} & & \textcolor[RGB]{205,24,24}{ {$\mathbf{0.311}$}} & \textcolor[RGB]{205,24,24}{ {$\mathbf{0.182}$}} & \textcolor[RGB]{205,24,24}{ {$\mathbf{0.123}$}} & $0.239$ &   \textcolor[RGB]{205,24,24}{ {$\mathbf{410.683}$}}  \\

        \bottomrule
    \end{tabular*}
    \begin{tablenotes}
            \footnotesize
        \item[1] '$-$' denotes that the parameter is not covered in the simulation.
        \item[2] Parameters learned through BO are highlighted in \textbf{bold}.
        \item[3] The best RMSE of each vehicle state is highlighted in \textcolor[RGB]{205,24,24}{\textbf{red bold}}.
    \end{tablenotes}
\end{threeparttable}
\end{table*}

It is observed that the proposed ALMPC can achieve rapid-convergent control performance for unbiased lateral error (Fig. \ref{case2_e}) and course error (Fig. \ref{case2_zeta}) among all three simulations.
The maximize lateral deviations are respectively $-1.46$ m for MPC-PPT results, $1.27$ m for MPC-APT, $1.32$ m for MPC-DEP,  and $0.51$ m for ALMPC, which also shows a significant lane-keeping ability of system learning.
Fig. \ref{case2_drifting_X} proves that the vehicle achieves enhanced performance under ALMPC control compared with other simulations, particularly in states such as  $V$, $\beta$, and $\delta$. 
It competes closely with the best-performing simulations in states of $r$ and $F_{xr}$, demonstrating that ALMPC has effectively addressed the conflicting control goals of path tracking and drifting.

With the learned APT control law and DEP from BO, the real vehicle path is close to the reference clothoid-based path while maintaining steady drift states. 
The performance cost $J(\theta)$ of ALMPC during the learning process is shown in Fig. \ref{case2_BO_iterations} and the best parameters are acquired in the 171 iterations.

\begin{figure}[!b]
\centering
\subfloat[Lateral error]{\includegraphics[width=1\linewidth]{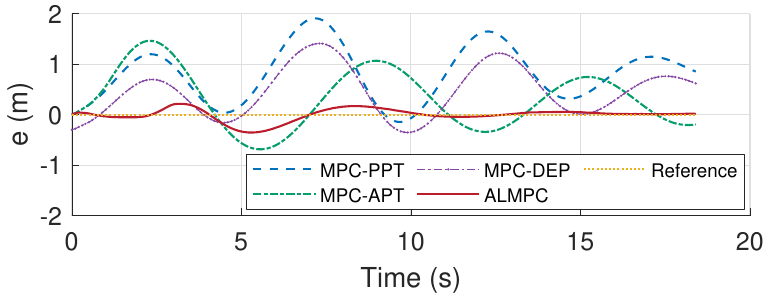}%
    \label{case4_e}}
\hfil
\subfloat[Course error]{\includegraphics[width=1\linewidth]{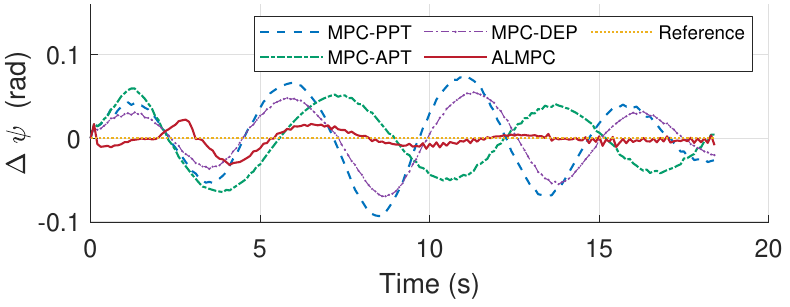}%
    \label{case4_zeta}}
\caption{Simulations results of the path tracking performance in case 2: the misidentified road friction coefficient triggers the vehicle to undulate around the reference path in simulations except the ALMPC, which can still maintain remarkable path tracking ability. }
\label{case4_tracking}
\end{figure}

\subsubsection{Case 2: Misidentified Vehicle Parameter}

To further validate the effectiveness of system learning through BO, misidentified parameters are deliberately utilized to aggravate the modeling error.
We assume a reduced road friction coefficient of $\mu_m = 0.9$ (a 10\% decrease of $\mu$) in Carsim to simulate the conditions of a slippery road surface that might occur with rain or snow. 
However, we continue to apply $\mu = 1$ in the Matlab platform to establish the precondition of biased modeling.

\begin{figure}[!ht]
\centering
\subfloat[Velocity]{\includegraphics[width=1\linewidth]{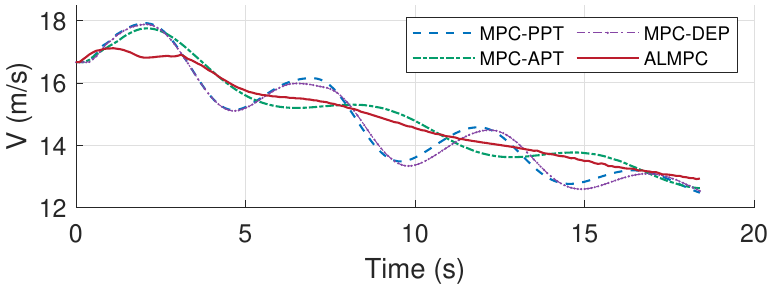}%
    \label{case4_v}}
\hfil
\subfloat[Sideslip angle]{\includegraphics[width=1\linewidth]{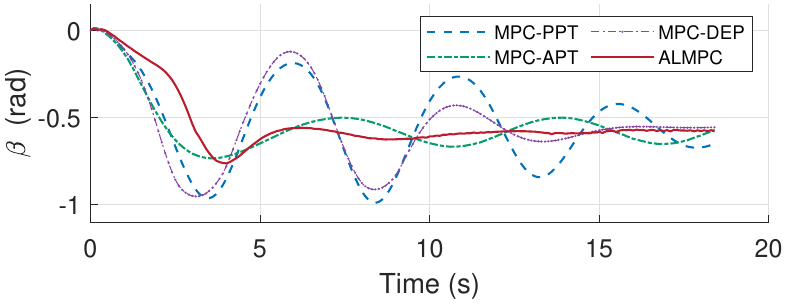}%
    \label{case4_beta}}
\hfil
\subfloat[Yaw rate]{\includegraphics[width=1\linewidth]{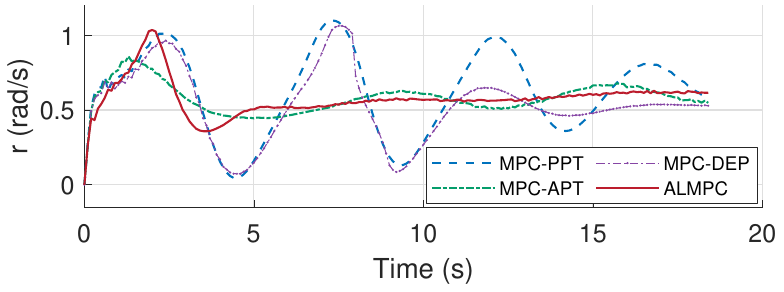}%
    \label{case4_r}}

\hfil
\centering
\subfloat[Steering angle]{\includegraphics[width=1\linewidth]{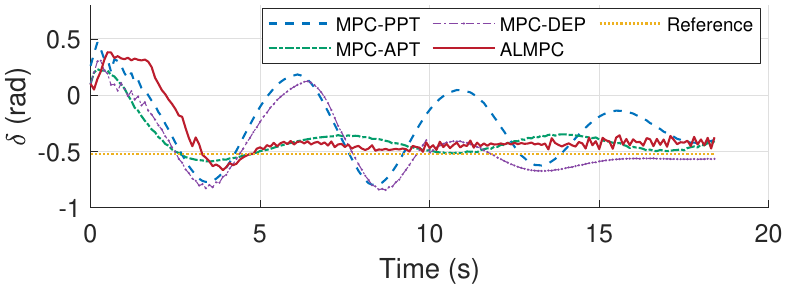}%
    \label{case4_delta}}
\hfil
\subfloat[Rear longitudinal force]{\includegraphics[width=1\linewidth]{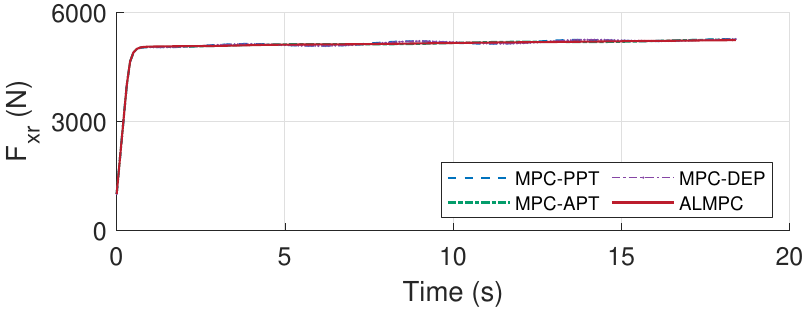}%
    \label{case4_F}}

\caption{Simulations results of drift vehicle states in case 2: the drifting performance of the proposed ALMPC proves to be less affected by the misidentified road friction coefficient than other simulations.} 

\label{case4_drifting_X}
\end{figure}

The system parameters and performance of three simulations are demonstrated in Table \ref{result2}.
The path-tracking performance of the vehicle is shown in Fig. \ref{case4_tracking} and the drift vehicle states are shown in Fig. \ref{case4_drifting_X}.
The superior performance of the proposed ALMPC becomes more evident under the condition of misidentified road friction coefficient, which is the only one achieving stable control performance for unbiased lateral error (Fig. \ref{case4_e}) and course error (Fig. \ref{case4_zeta}) among all three simulations.

\begin{figure}[!t]
	\centering
	\includegraphics[width=0.9\linewidth]{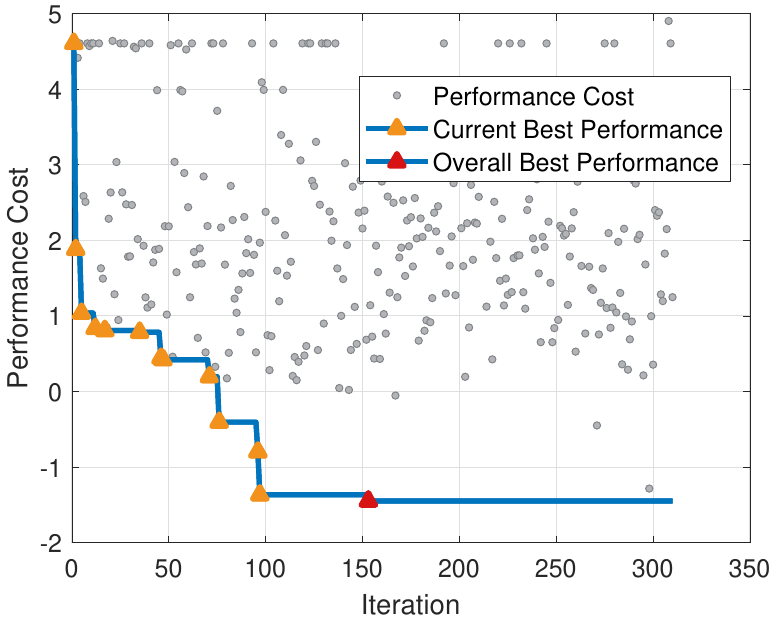}
	\caption{The path tracking performance cost during BO process in case 2: the BO learning process can significantly improve the path tracking performance and the optimal set of parameters is acquired in iteration 153.  }
	\label{case4_BO_iterations}
\end{figure}

The maximize lateral deviations are respectively $1.91$ m for MPC-PPT results, $1.46$ m for MPC-APT, and $0.21$ m for ALMPC, which indicate the prominent lane-keeping ability of system learning.
The vehicle can also achieve enhanced drifting performance under ALMPC control compared with other simulations, demonstrating that ALMPC has effectively addressed the modeling error and conflicting control goals.

With the learned APT control law and the DEP from BO, the vehicle can demonstrate significant path tracking and drifting performance even with the misidentified road friction coefficient.
The performance cost $J(\theta)$ of ALMPC during the learning process is shown in Fig. \ref{case4_BO_iterations} and the best parameters are acquired in the 153 iteration.

\subsection{Discussion}
The verification of the proposed ALMPC strategy encompasses two pivotal aspects relating to learning the APT control law and the DEP, each contributing uniquely to improving vehicle performance. 

\subsubsection{The Proposed ALMPC Strategy }
\textbf{\textit{APT Learning:}} The proposed APT control law enhances path tracking performance without compromising drifting performance, which can successfully balance the control conflict and even achieving improvements in both aspects. 
However, existing controller oscillations demonstrate the restricted performance improvements due to the mismatched system model, which underlines the necessity of learning DEP to compensate for modeling errors.

\textbf{\textit{DEP Learning:}} The learned DEP can effectively compensate for uncertainties in the system model, making it suitable to address real-world challenges like environmental variations and parameter mismatches.
However, the persistent steady-state errors highlights the need for an adaptive mechanism, such as the APT control law, to further enhance path tracking performance while maintaining robust drifting.

\textbf{\textit{APT \& DEP Learning:}} The previous analysis reveals that optimizing a single aspect is insufficient due to the coupled relationship between path tracking and drifting, emphasizing the need for a unified learning framework. 
The proposed ALMPC strategy, integrating both the APT control law and DEP, demonstrates significant reductions in root mean square errors across nearly all states. 
These results underscore the effectiveness of the proposed approach in handling complex system dynamics under challenging conditions.

\subsubsection{ An 8-shaped Drift Maneuver }
\begin{figure}[!h]
	\centering
	\includegraphics[width=1\linewidth]{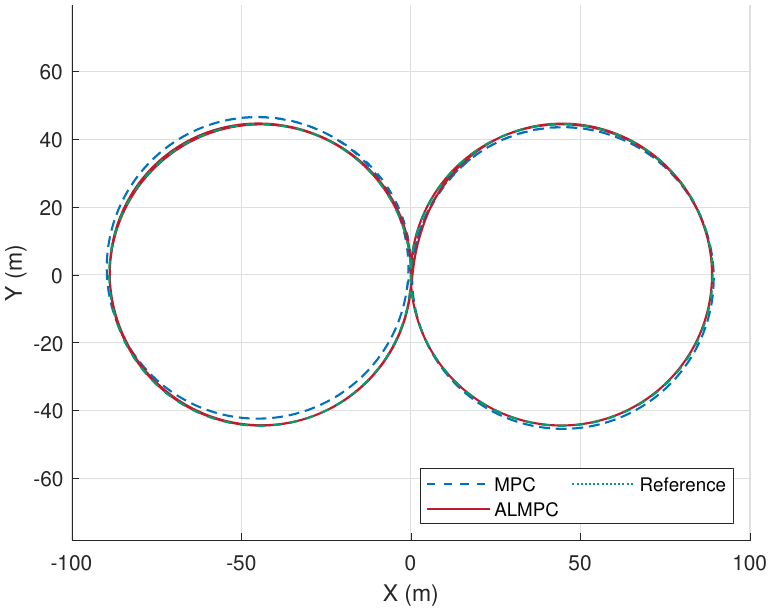}
	\caption{8-shaped Drifting Simulation: The drift vehicle controlled by ALMPC remains closely aligned with the reference path throughout the 8-shaped path, whereas the traditional MPC exhibits noticeable deviations from the reference path.   }
	\label{8shape_position}
\end{figure}

To prove the generalization of the proposed ALMPC strategy, we conduct an additional simulation of an 8-shaped drift maneuver, which requires frequent drift switching and a high level of steering adaptability.
The 8-shaped drifting simulation is illustrated in Fig. \ref{8shape_position}, which compares the performance between the ALMPC and a traditional MPC without the ALMPC strategy.  
Although the traditional MPC can achieve similar drift performance, it is not as effective as the ALMPC due to existing modeling errors. 
Additionally, it fails to follow the given path, which could lead to unsafe driving behaviors. This issue arises from the conflicting control objectives of path tracking and drifting, which are difficult for traditional controllers to reconcile.

\subsubsection{ Comparison with Deep Reinforcement Learning }

\begin{figure}[!b]
	\centering
	\includegraphics[width=1\linewidth]{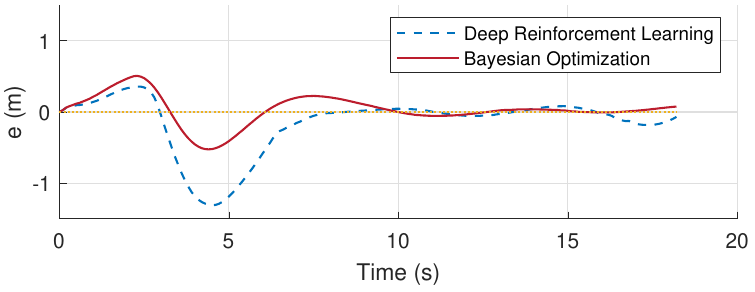}
	\caption{Path tracking comparison between BO and DRL: DRL exhibits more oscillations compared to BO, with a maximum lateral error of -1.31 m, while BO only 0.51 m. }
	\label{BO2RL_e}
\end{figure}

The model-free deep reinforcement learning (DRL)-based controller presented in \cite{HighSpeedAutonomous2020a} relies on a well-designed training process and experience-based drift-driving data as initial references to promise effective learning, which can be challenging to acquire for drift vehicles. 
This can restrict the generalization ability of DRL-based methods to new drifting scenarios. 
We design the DRL framework with the DEP derived from system model, which can promote safe drifting behaviors especially in the early stages of training, to serve as a comparison with our BO-based approach.
As illustrated in Fig. \ref{BO2RL_e}, the proposed ALMPC strategy based on BO exhibits fewer oscillations compared to DRL and a 10 \% less RMSE for the steering angle.
This swift online execution and shorter offline training time illustrated in Table \ref{BO2RL} also highlight the efficiency of the proposed BO-ALMPC approach, making it highly suitable for real-time driving applications.

\begin{table}
\caption{ Computational Complexity Evaluation
\label{BO2RL}}
\centering
\begin{tabularx}{\tblwidth}{l|X X}

    \toprule
    & Approach  & Time \\
    \midrule
\multirow{2}{*}{ \textbf{Offline Training}  } 
            & BO-ALMPC  & \SI{0}{\hour} \SI{37}{\minute}\\
            & DRL-ALMPC &  \SI{12}{\hour} \SI{17}{\minute} \\
    \midrule
\multirow{2}{*}{ \textbf{Online Execution} } 
            & BO-ALMPC & \SI{9.32}{\second} \\
            & DRL-ALMPC &  \SI{26.85}{\second} \\

    \bottomrule
\end{tabularx}

\end{table}

\subsubsection{ Comparison with Other Search Strategies }
\begin{figure}[!t]
	\centering
	\includegraphics[width=1\linewidth]{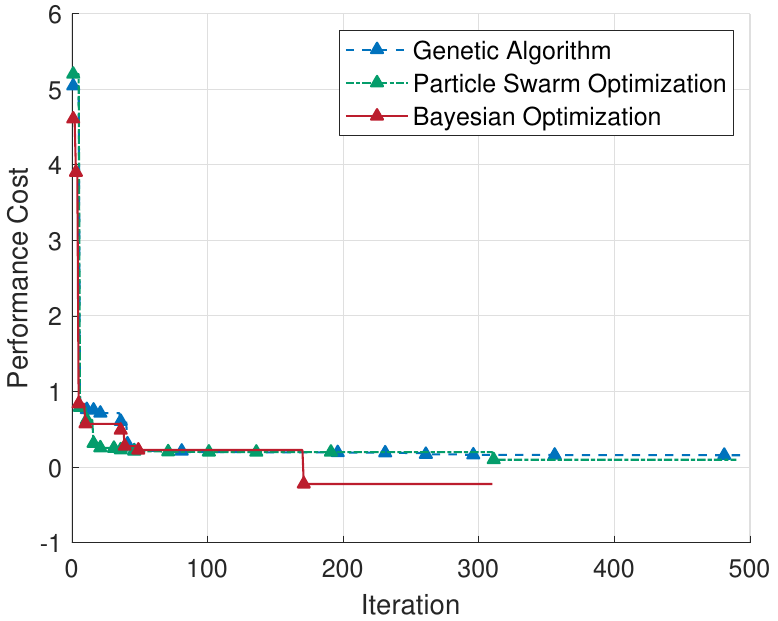}
	\caption{Comparison among three optimization approaches: BO can achieve superior optimization results with fewer evaluations. }
	\label{compare_iterations}
\end{figure}

Gradient-based methods are not applicable in this context due to the difficulty in analytically modeling the relationship between the parameter set and the path tracking performance.
While evolutionary strategies, such as genetic algorithms and particle swarm optimization, often require a significantly larger number of evaluations during each iteration to converge to an optimal solution \cite{PerformanceDrivenCascade2022}, making them less effective for the expensive-to-evaluate drifting simulations.
We compare the optimization process of these two evolutionary strategies with BO as illustrated in Fig. \ref{compare_iterations}
Simulation results indicate that BO can achieve superior optimization results with fewer evaluations, highlighting its suitability for drifting scenarios.

\section{Conclusion}
This paper presents an ALMPC strategy to handle the complicated drift dynamics and uncertain driving environments.
The ALMPC strategy employs a hierarchical structure: an upper-level BO supervisor learning the DEP and APT control law and a lower-level MPC drift controller, which can effectively decouple the control conflicts between path tracking and drifting.
Simulation results on the Matlab-Carsim platform demonstrate that the proposed ALMPC strategy can achieve stable drifting performance, while safely adhering to the desired clothoid-based path even under the scenario with misidentified vehicle parameters. 

In future work, we plan to implement our approach on the F1Tenth platform, which closely aligns with the low computational requirements of our controller design. This can further evaluate the model learning ability in the DEP identification process under complex physical conditions, as well as the effectiveness of the APT control law across different reference paths. This will provide a deeper understanding of the adaptability and robustness of the ALMPC strategy in real-world driving scenarios.

\bibliographystyle{unsrt}
\bibliography{journal_reference}

\newpage
\bio{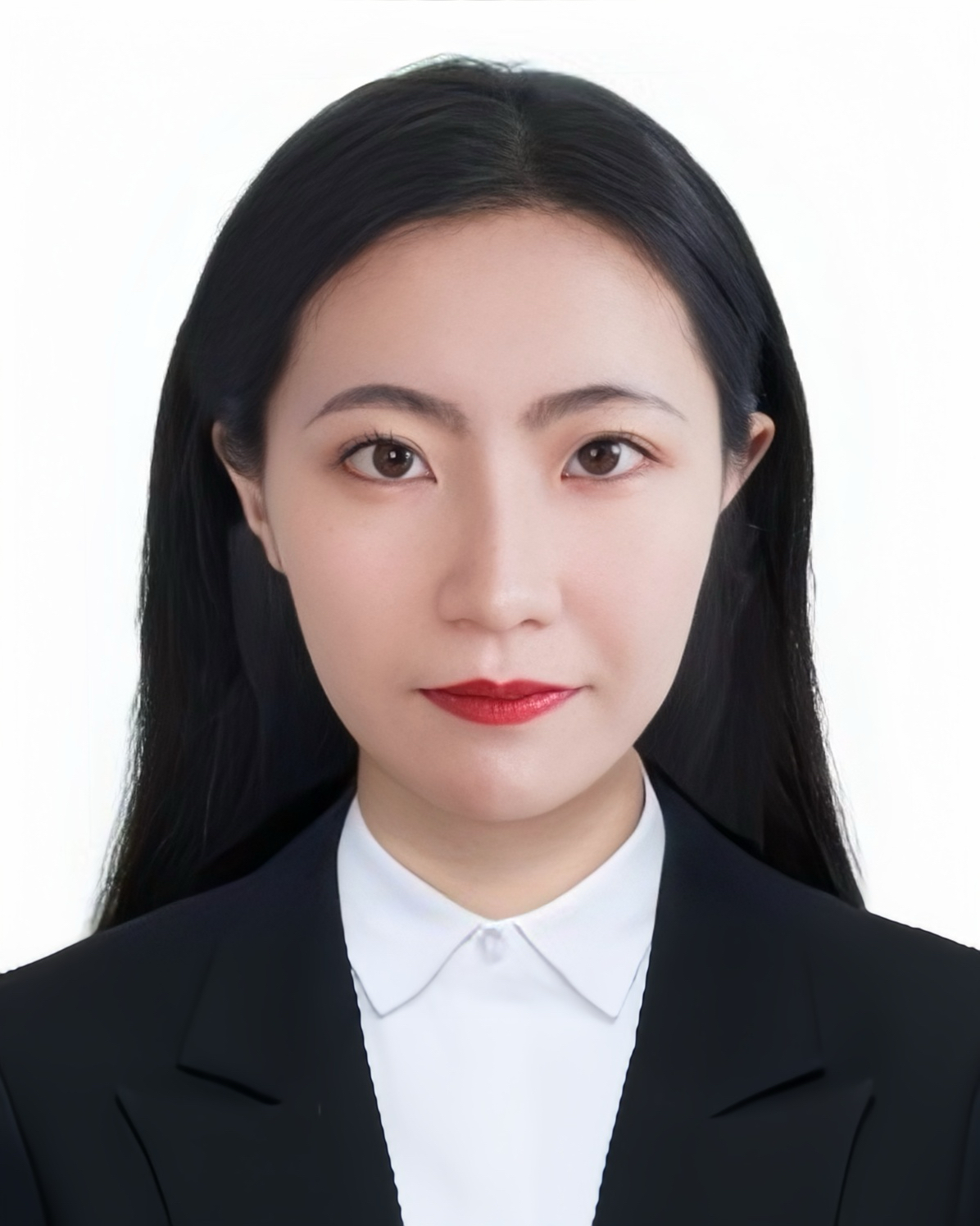}
{Bei Zhou} received the B.S. degree in communication engineering from Beijing Institute of Technology, Beijing, China, in 2021. She is currently working towards the Ph.D. degree at the State Key Laboratory of Industrial Control Technology, Institute of Cyber-Systems and Control, Zhejiang University, Hangzhou, China.
Her main research interests include vehicle dynamics control, motion planning and path tracking.
\endbio

\vspace{15 mm} 
\bio{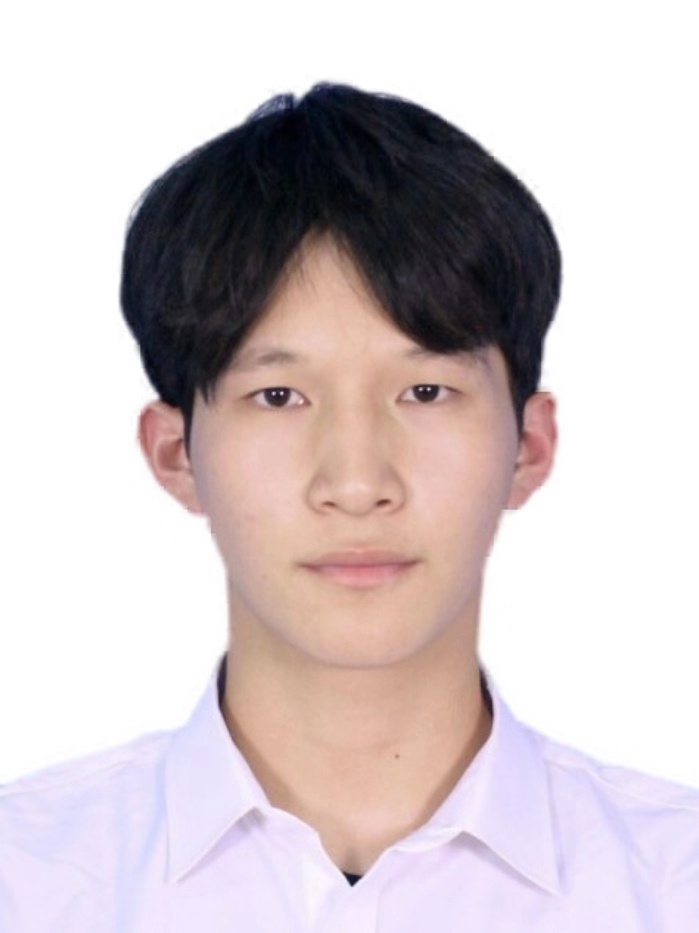}
{Cheng Hu} received the B.S. degree from Nanjing Agricultural University, NanJing, China, in 2020. He is currently pursuing the Ph.D. degree with the College of Control Science and Engineering in Zhejiang University, Hangzhou, China. His research interests include aggressive driving, path planning, model predictive control and machine learning.
\endbio

\vspace{20 mm} 
\bio{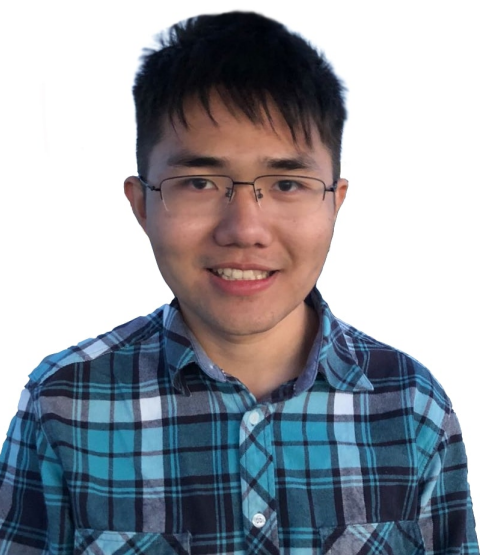}
{Jun Zeng} received his Ph.D. in Control and Robotics at the Department of Mechanisl Engineering at University of California, Berkeley, USA in 2022 and Dipl. Ing. from Ecole Polytechnique, France in 2017, and a B.S.E degree from Shanghai
Jiao Tong University (SJTU), China in 2016. His research interests lie at the intersection of optimization, control, planning, and learning with applications on various robotics platforms.
\endbio

\vspace{20 mm} 
\bio{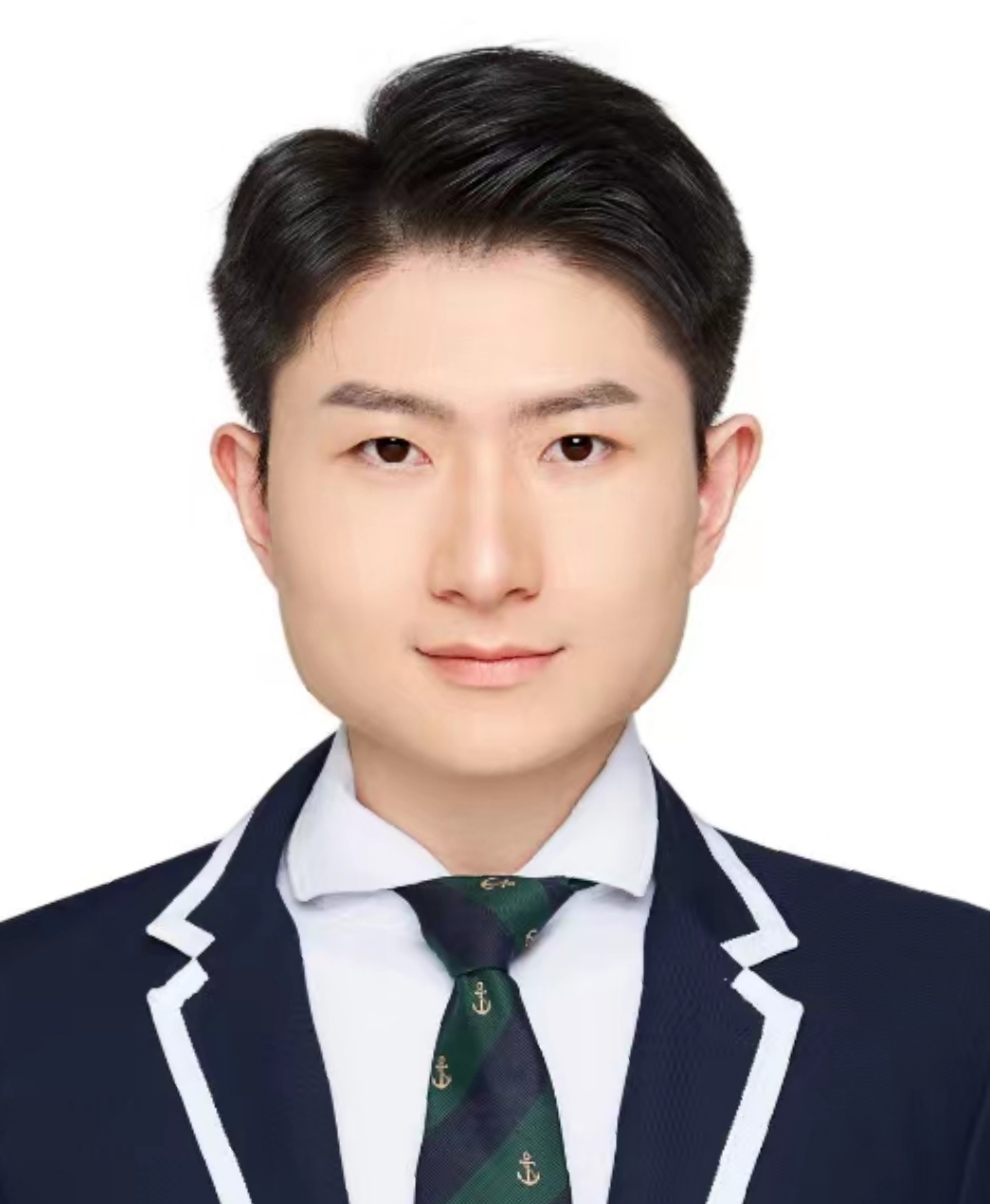}
{Zhouheng Li} is currently working toward the Ph.D. degree at the College of Control Science and Engineering, Zhejiang University, Hangzhou, China. His research centers on developing trajectory planning methods for aggressive autonomous driving, with applications in high-speed racing and drifting. He is also actively involved in applying these techniques in the F1TENTH competition. His broader research interests include behavior planning, deep reinforcement learning, and Bayesian optimization.
\endbio

\newpage
\bio{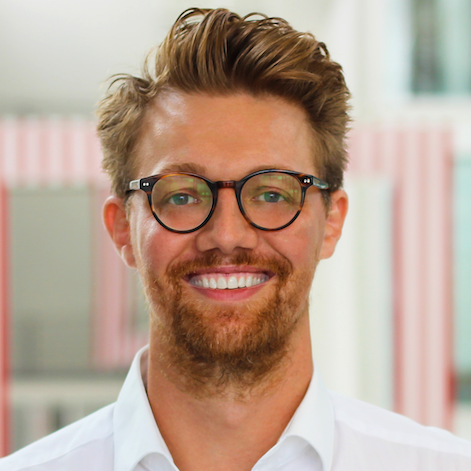}
{Johannes Betz}~is an assistant professor in the Department of Mobility Systems Engineering at the Technical University of Munich (TUM), where he is leading the Autonomous Vehicle Systems (AVS) lab. He is one of the founders of the TUM Autonomous Motorsport team. His research focuses on developing adaptive dynamic path planning and control algorithms, decision-making algorithms that work under high uncertainty in multi-agent environments, and validating the algorithms on real-world robotic systems. Johannes earned a B.Eng. (2011) from the University of Applied Science Coburg, a M.Sc. (2012) from the University of Bayreuth, an M.A. (2021) in philosophy from TUM, and a Ph.D. (2019) from TUM.
\endbio

\vspace{10 mm} 
\bio{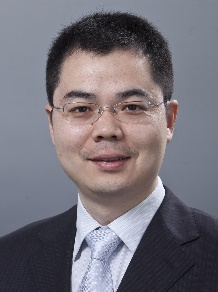}
{Lei Xie} received a B.S. degree in 2000 and a Ph.D. in 2005 from Zhejiang University, P.R. China. Between 2005 and 2006, he was a postdoctoral researcher at Berlin University of Technology, an Assistant Professor between 2005 and 2008 and is currently a Professor at the Department of Control Science and Engineering, Zhejiang University. To date, his research activities culminated in over 30 articles that are published in internationally renowned journals and conferences, 3 book chapters and a book in the area of applied multivariate statistics and modeling. His research interests focus on the interdisciplinary area of statistics and system control theory.
\endbio

\vspace{10 mm} 
\bio{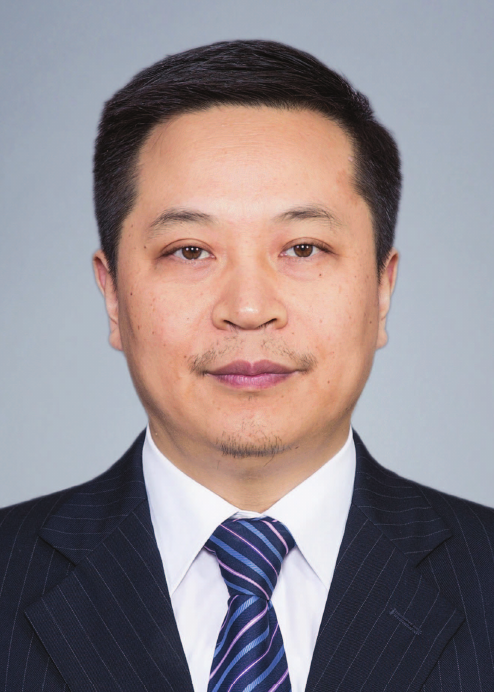}
{Hongye Su}  was born in 1969. He received the B.S. degree in industrial automation from the Nanjing University of Chemical Technology, Jiangsu, China, in 1990 and the M.S. and Ph.D. degrees from Zhejiang University, Hangzhou, China, in 1993 and 1995, respectively. He was a Lecturer with the Department of Chemical Engineering, Zhejiang University from 1995 to 1997, where he was an Associate Professor with the Institute of Advanced Process Control from 1998 to 2000, and currently a Professor with the Institute of Cyber-Systems and Control. His current research interests include the robust control, time-delay systems, and advanced process control theory and applications.
\endbio

\end{document}